%% file: iclr2027_conference.tex
\documentclass{article}
\usepackage{iclr2027_conference,times}
\usepackage[T1]{fontenc}
\usepackage[utf8]{inputenc}
\input{math_commands.tex}

\usepackage{hyperref,url,graphicx,booktabs,multirow,xcolor,microtype}
\usepackage{pifont}
\newcommand{\cmark}{\ding{51}}
\newcommand{\xmark}{\ding{55}}
\usepackage{caption}
\usepackage{wrapfig}
\usepackage{enumitem}
\usepackage{xparse}
\newcommand{\method}{RSIMed}
\title{\method{}: Scaling Medical Foundation Models\\beyond Limited Clinical Data\\via Retrieval-Grounded Self-Evolution}
\author{Wei Dai$^{1}$, Qianchu Liu$^{2}$, Guanghui Qin$^{2}$, Xing Han$^{3}$, Tristan Naumann$^{2}$, \\\textbf{Sheng Zhang$^{2}$, Paul Pu Liang$^{1}$} \\
$^{1}$ Massachusetts Institute of Technology, $^{2}$ Microsoft Research, $^{3}$ Johns Hopkins University}
\arxivpreprintcopy
\begin{document}
\raggedbottom
\maketitle
\begin{center}
\vspace{-1em}
\includegraphics[width=\textwidth]{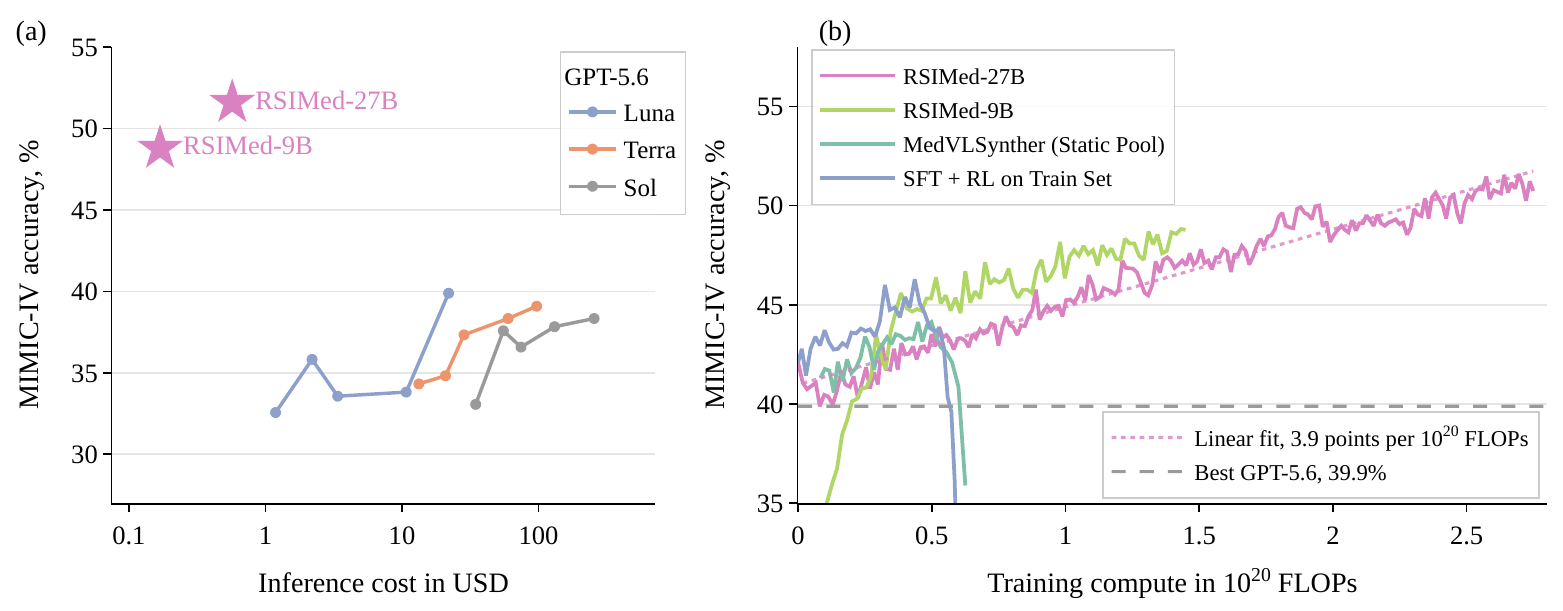}
\vspace{-1.5em}
\captionof{figure}{\textbf{RSIMed continues improving with training compute and surpasses frontier models on rare-disease diagnosis.} \textbf{(a)} RSIMed-27B and RSIMed-9B outperform off-the-shelf GPT-5.6 Luna, Terra, and Sol on MIMIC-IV at every reasoning effort, for a small fraction of the inference cost over the 2,452 evaluation cases. \textbf{(b)} The accuracy of RSIMed-27B grows linearly with training compute, while RL on the labeled training set and RL on a static pool written once by the same proposer, the fixed-corpus setting of MedVLSynther, both peak and then collapse. In our runs, $10^{20}$ FLOPs corresponds to about 325 B200 GPU hours of training. }
\label{fig:teaser}
\end{center}
\begin{abstract}
Rare diseases are hard to diagnose, and the labeled clinical records needed to train models for them are few, private, and slow to extend, since a specialist must read every case. A model that writes its own training tasks could learn beyond these records, but no program can check a model-written diagnosis, and a wrong answer key corrupts training. To tackle this dilemma between data scarcity and difficulty of data annotations, we introduce \textbf{RSIMed}, \textbf{R}ecursive \textbf{S}elf-\textbf{I}mprovement in \textbf{Med}icine, one of the first self-evolving pipelines in which a task proposer co-evolves in context with a diagnostic solver on real clinical records. Seeded by training admissions, a frozen copy of the solver writes each clinical question from retrieved medical evidence, so its answer rests on a source, and steers topics and difficulty toward what the solver still gets wrong. On rare-disease diagnosis from MIMIC-IV admissions, the accuracy of RSIMed-27B grows linearly with training compute, while a static pool of generated tasks collapses. Without any external model, RSIMed-27B and RSIMed-9B reach 0.516 and 0.488 diagnostic accuracy, above every off-the-shelf GPT-5.6 setting, whose best reaches 0.399 at 38 times the inference cost of RSIMed-27B. Tasks written by four external teachers, including GPT-5.5, train a less accurate diagnostician than the solver's own. Trained only on diagnosis, RSIMed-RL-27B also improves on the physician-written tasks of HealthBench Professional, from 0.291 to 0.303 in adjusted score.
\end{abstract}

\input{figures/main_pipeline}
\input{sections/introduction}
\input{tables/kb_sources}
\input{sections/related_work}
\input{sections/method}
\input{sections/experiments}
\input{sections/conclusion}

\clearpage
\input{sections/statements}
\bibliography{ref}
\bibliographystyle{iclr2027_conference}

\clearpage
\appendix

\input{sections/appendix_related}

\input{sections/appendix_pipeline}
\input{sections/appendix_generated_tasks}

\input{sections/appendix_training}
\input{sections/appendix_baselines}
\input{sections/appendix_judging}

\input{sections/appendix_results}
\input{sections/appendix_teacher_analysis}
\clearpage
\input{sections/appendix_examples}

\input{sections/appendix_extended_results}

\input{sections/appendix_limitations}
\input{sections/appendix_availability}
\end{document}

%% file: math_commands.tex
\usepackage{amsmath,amsfonts,bm}

\def\eqref#1{equation~\ref{#1}}

\def\1{\bm{1}}

\DeclareMathAlphabet{\mathsfit}{\encodingdefault}{\sfdefault}{m}{sl}
\SetMathAlphabet{\mathsfit}{bold}{\encodingdefault}{\sfdefault}{bx}{n}

%% file: figures/main_pipeline.tex
\begin{figure*}[t]
\centering
\includegraphics[width=\textwidth]{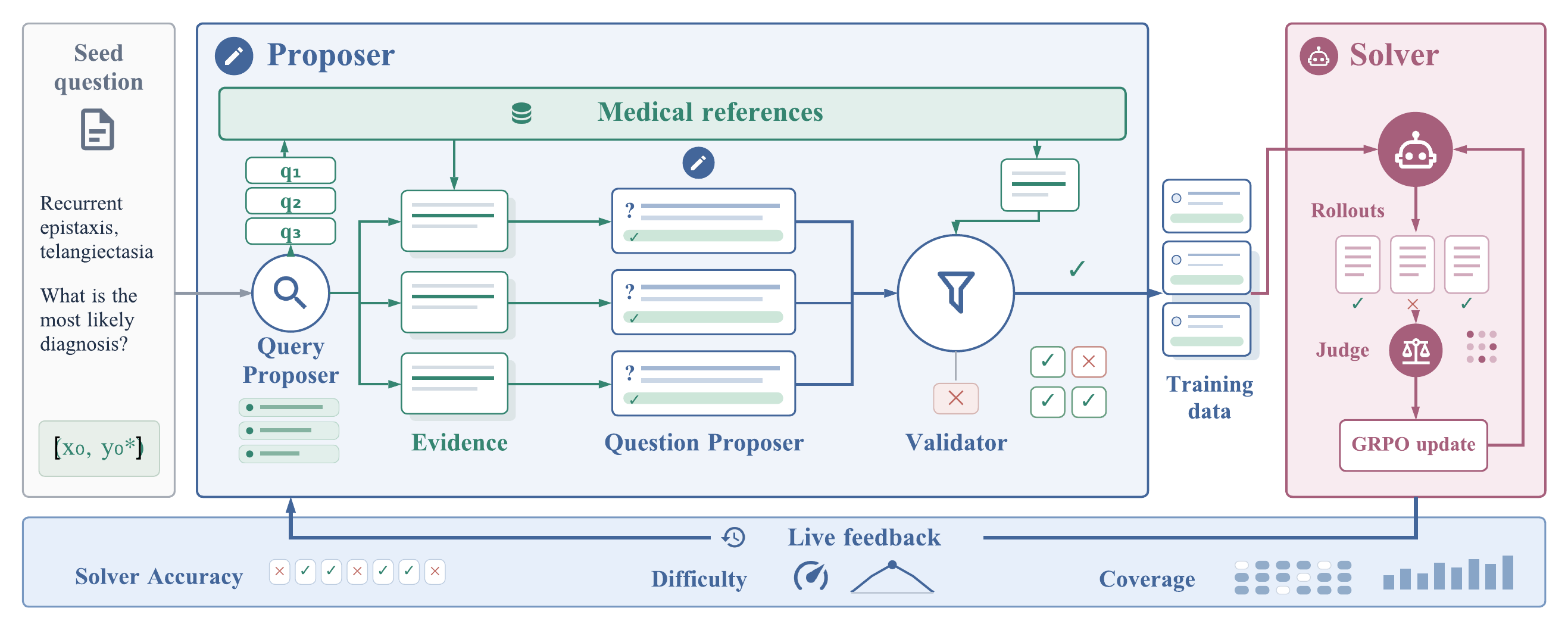}
\vspace{-1.5em}
\caption{\textbf{The \method{} training pipeline.} Starting from a clinical seed question, the query proposer retrieves medical evidence and the question proposer generates questions with reference answers. A validator checks format and rejects incoherent questions or answers contradicted by newly retrieved evidence. Accepted tasks enter the solver's pool. Multiple responses receive judge-based rewards for GRPO updates. Solver accuracy and task history guide subsequent difficulty and coverage.}
\label{fig:pipeline}
\end{figure*}

%% file: sections/introduction.tex
\section{Introduction}
Foundation models have advanced quickly in general reasoning and in medicine \citep{yang2025qwen3,singhal2025toward,sellergren2025medgemma}, but diagnosing rare diseases from clinical records remains difficult. Labeled cases are scarce, because most clinical records are private and a specialist must read each record to confirm its diagnosis. Recursive Self-Improvement (RSI) methods, in which a model generates the supervision it learns from, can reduce the need for such labels. A model can bootstrap its own rationales, reward its own outputs, or pair a proposer that writes tasks with a solver that learns them \citep{zelikman2022star,yuan2024self,chen2024self,zhao2026absolute,huang2026r,chen2025self}. These methods have driven progress in mathematics, coding, and scientific discovery \citep{gao2025survey,qu2026coral,novikov2025alphaevolve,zhang2026darwin}, where a program, a unit test, or a final number checks every answer. However, none of these automatic methods works in medicine.

A model that writes its own clinical training tasks could train far beyond the labeled cases, but how to generate tasks that stay correct and keep teaching the solver, without a stronger teacher, remains the core unsolved problem. Medical models have been trained on generated problems in various forms \citep{zhang2024ultramedical,chen2024huatuogpt,wu2025medreason,huang2026synthesizing}. However, these methods fails to adapt to the dynamically changing capability of the model. Methods usually build its training data once, before training, so its tasks become too easy as the solver improves, and repeated training on generated data can reduce accuracy \citep{shumailov2023curse}. In addition, a generated clinical task has no correctness check. No program can catch an incorrect clinical key, and wrong generated labels reinforce the model's own errors \citep{liu2026self}.

To address these challenges, we propose \textbf{RSIMed}, \textbf{R}ecursive \textbf{S}elf-\textbf{I}mprovement in \textbf{Med}icine, one of the first self-evolving pipelines where the proposer co-evolves with the solver in context, on free-text diagnosis from real clinical records. To adapt to the changing capability of the solver, we propose a live feedback framework that dynamically calibrates the questions based on the solver's past performance. To address the verification problems, we propose a validator that checks generated problems against our collected medical database. We show that with this pipeline, the model is able to improve itself without the need of a stronger teacher model.

In this work, our contributions are as follows.
\begin{enumerate}[leftmargin=*]
\item We propose RSIMed, a self-evolving pipeline in which a frozen proposer co-evolves in context with a solver trained by reinforcement learning. The proposer writes each task from retrieved evidence and steers its difficulty and coverage by the solver's accuracy, a validator rejects tasks that fresh evidence contradicts, and a judge grades free-text diagnoses semantically.
\item We evaluate RSIMed in and out of domain. On rare-disease diagnosis from MIMIC-IV, RSIMed-27B accuracy grows linearly with training compute, by 3.9 points per $10^{20}$ FLOPs, while a static pool of generated tasks collapses. RL lifts RSIMed-9B and RSIMed-27B from 0.288 and 0.421 after the supervised warm start to 0.488 and 0.516, above every off-the-shelf GPT-5.6 configuration at a fraction of its cost. Trained only on diagnosis, RSIMed-RL-27B also raises its adjusted HealthBench Professional score from 0.291 to 0.303.
\item We analyze where the learning signal of generated tasks comes from. A stronger teacher does not always produce a stronger student, as 36\% of the solver's own short-answer tasks yield rollout groups with mixed correctness against 22\% of GPT-5.5 tasks, and an independent audit rates 84\% of accepted tasks as valid.
\end{enumerate}
We release all code, prompts, and configurations at \url{https://github.com/DDVD233/RRI}.

%% file: tables/kb_sources.tex
\begin{table*}[t]
\centering\footnotesize
\caption{Sources of the medical knowledge base. Entries and words count the indexed text that a
retrieval returns. Img./vid. counts the distinct image and video files that the multimodal entries
link to. These entries are retrieved by their text and bring their image or video with them.}
\vspace{-0.5em}
\label{tab:kb-sources}
\scriptsize\setlength{\tabcolsep}{3pt}
\begin{tabular*}{\textwidth}{@{\extracolsep{\fill}}llrrr@{}}
\toprule
Source & Content & Entries & Words & Img./vid. \\
\midrule
PubMed abstracts & Passages of PubMed abstracts from clinical trials, case reports, and reviews & 23{,}898{,}701 & 1.74B & -- \\
Wikipedia & Passages of English Wikipedia articles on medicine and general topics & 29{,}913{,}202 & 2.20B & -- \\
Medical textbooks & Passages of 18 medical textbooks, from anatomy and pathology to surgery & 125{,}806 & 8.9M & -- \\
WikiDoc & Clinician-written medical encyclopedia, mostly titles with a few full articles & 190{,}530 & 992.8K & -- \\
PMC-VQA & Question-answer pairs on figures from PubMed Central \citep{zhang2023pmc} & 226{,}943 & 3.2M & 149{,}073 \\
MIRAGE & Medical exam and biomedical research questions \citep{xiong2024benchmarking} & 5{,}120 & 78.3K & -- \\
PubMedQA & Yes, no, or maybe questions, each with its PubMed abstract \citep{jin2019pubmedqa} & 512 & 37.2K & -- \\
CLIMB & Diagnostic questions on clinical images and videos \citep{dai2025climb} & 2{,}619{,}312 & 107.3M & 2{,}538{,}949 \\
\midrule
Total & & 56{,}980{,}126 & 4.06B & 2{,}688{,}022 \\
\bottomrule
\end{tabular*}
\vspace{-1em}
\end{table*}

%% file: sections/related_work.tex
\section{Related Work}
\label{sec:related}

\textbf{Medical models and their training data.} Medical models range from prompted general models to specialized and reinforcement-learned ones \citep{singhal2023large,sellergren2025medgemma,xu2025lingshu,wang2025baichuan,chen2024huatuogpt,lai2026med,huang2025m1,zhang2025med}, and we build on the clinical setting of QoQ-Med \citep{dai2026qoq}. Many are trained on data written by a stronger model \citep{zhang2024ultramedical,wu2025medreason,yilmaz2025synthetic}. In general domains, synthetic corpora written by a teacher or generated from textbooks and documents \citep{mukherjee2023orca,gunasekar2023textbooks,abdin2024phi,yehudai2024genie,yang2025synthetic} can favor style over capability and degrade without real data \citep{gudibande2023false,shumailov2023curse,gerstgrasser2024model,feng2025beyond}. EDCO reorders existing examples during training \citep{pang2026edcodynamiccurriculumorchestration}, and medical agents update their prompts at inference \citep{almansoori2025medagentsim,liu2026evomdt}. All of these methods depend on a stronger teacher or on data fixed before training.

\textbf{Self-improvement and self-play.} Self-improvement trains a model on its own verified or self-judged outputs \citep{zelikman2022star,gulcehre2023reinforced,yuan2024self,chen2024self,huang2025self}, and proposer--solver and self-play methods also write the tasks \citep{chen2025self,wilf2025propose,zhou2026self,huang2026g,sui2026conversation}. To label the tasks they write, Absolute Zero runs a program \citep{zhao2026absolute}, R-Zero takes a majority vote \citep{huang2026r}, SPICE extracts answer keys from documents \citep{liu2025spice}, and WIST brings a challenger--solver loop with web retrieval to medicine but grades multiple-choice exams with rules \citep{li-etal-2026-wist}. None of these checks confirms a free-text diagnosis, and wrong labels reinforce errors in medicine \citep{liu2026self}. RSIMed is the only method in Table~\ref{tab:related} that both generates tasks throughout training and checks free-text answers against evidence.

%% file: sections/method.tex
\section{Method}
\label{sec:method}
\label{app:extended-method}

The core challenge is data scarcity, which rephrasing database entries cannot solve. We therefore design RSIMed to maximize data diversity at the knowledge frontier, even when the correctness of an answer is uncertain. As shown in Figure~\ref{fig:pipeline}, RSIMed consists of four components. A proposer writes questions and reference answers from retrieved medical evidence, and a validator checks each task against fresh evidence, because no program can verify a clinical answer. A solver learns from accepted tasks without seeing the evidence, and a judge grades its responses by meaning, because one diagnosis has many phrasings. The solver co-evolves with the proposer, the solver in its weights through reinforcement learning and the frozen proposer in context from its solve rates. In the self-improvement setting, a frozen copy of the solver's base model fills every other role, and in the teacher-student setting, an external teacher writes the tasks. Below we describe the details of each component.

\input{figures/task_example}

\subsection{Problem Definition}
\label{sec:problem}

Our goal is to improve a solver $\pi_\theta$ on a clinical task, here rare-disease diagnosis,
from a limited set of labeled cases. Given a clinical question $q$, the solver produces a
response $y$ with reasoning and a final answer $\hat{y}$. Each training task pairs a question
with a reference answer $y^\ast$. Besides the labeled cases, at training step $t$ the proposer
$P$ writes a multiple-choice or short-answer task $(q,y^\ast)\sim P(\cdot\mid s,E_s,H_t)$ from
a seed case $s$ of the training split, the passages $E_s$ retrieved for it, and a feedback
history $H_t$ of how the solver with weights $\theta_t$ answered earlier tasks. The solver
trains on the pool of labeled and generated tasks. In other words, the solver changes its
weights, the proposer changes only what it reads, and the training data are no longer bounded
by the number of labeled cases.

\subsection{Proposer}
\label{sec:proposer}

No executable test checks a clinical reference answer, and the model's own belief may be
wrong. The proposer therefore writes every task from evidence retrieved from a medical
knowledge base, so each reference answer traces to a source passage that the validator can
check again. The knowledge base indexes 57 million passages and question-answer
pairs from the eight public medical sources of Table~\ref{tab:kb-sources}, linking 2.7 million
medical images and videos.

\textbf{Task generation.} The proposer uses training cases as seeds and writes questions about
the medical knowledge behind them. The query proposer writes diverse search queries for a seed,
such as its disease mechanisms, diagnostic criteria, and differential diagnoses, and retrieves
passages for each. The question proposer then writes a multiple-choice or open-ended question
and its reference answer from these passages, and the answer must follow from the evidence and
standard medical knowledge. Figure~\ref{fig:task-example} follows one task from its seed case to
validation, and Appendix~\ref{app:examples} gives more examples.

\textbf{In-context co-evolution.} The proposer learns from the solver through its prompt. For
coverage, it sees the ten most similar previous questions from the history of accepted tasks
with their solve rates, which steers it toward topics the solver has failed or not yet seen. For
difficulty, it sees the solver's recent accuracy and a target of $p^\ast=0.5$. Above the target,
it writes harder questions with finer clinical distinctions, and below it, it phrases questions
more precisely to remove ambiguity. In other words, the proposer aims for tasks that the solver
answers correctly about half of the time, where the variance $p(1-p)$ of the binary correctness
reward, and with it the learning signal, peaks. Figure~\ref{fig:seed-evolution} shows the tasks for one training case becoming more
specific across four visits. Generation runs
asynchronously while the solver trains, as detailed in Appendix~\ref{app:pipeline}.

\input{figures/base_vs_self}

\subsection{Validator}
\label{sec:validator}

The validator decides whether a generated task enters the training pool. It checks the format,
then retrieves new passages with the question and its options and rejects incoherent questions and answers that
these passages directly contradict. A task is accepted when the evidence is inconclusive. This
does not guarantee a correct answer, but it keeps the question pool as diverse as possible, lets
the proposer explore the undetermined frontier of knowledge beyond our database, and still
rejects the hard errors that would corrupt training. Accepted tasks join the labeled cases in
the solver's pool.

\subsection{Solver}
\label{sec:solver}

The solver is trained in two stages, supervised fine-tuning (SFT) followed by reinforcement
learning (RL). SFT teaches the solver to reason toward a single final diagnosis, with
traces that the base model writes for each labeled training case after receiving its reference
diagnosis. RL then carries learning beyond the labeled cases, on the task stream that the
proposer keeps adapting to the solver. We initialize from the SFT checkpoint and train with DAPO
\citep{yu2026dapo} and the group-relative advantages of GRPO
\citep{shao2024deepseekmathpushinglimitsmathematical}. For each sampled task $q$, we generate
$G=8$ responses with judge rewards $r_1,\ldots,r_G$ and compute
$\hat{A}_i = (r_i-\bar{r}_q)/(s_q+\epsilon)$ with $\bar{r}_q=\tfrac{1}{G}\sum_{j=1}^{G}r_j$,
where $s_q$ is the standard deviation of the rewards and $\epsilon$ ensures numerical stability.
Groups with equal rewards have zero advantage, which is why the proposer aims for intermediate
difficulty.

\subsection{Judge}
\label{sec:reward}

The judge grades every solver response against the reference answer. Exact matching alone is too strict for clinical answers. Following prior reinforcement
learning work that rewards exact correctness and output format \citep{shao2024deepseekmathpushinglimitsmathematical,lai2026med}, agreement with the reference judged by a language model \citep{chen2024huatuogpt,su2025crossing,ma2026general}, graded scores against a reference \citep{gunjal2026rubrics}, reasoning quality \citep{fan2026sophiavl}, and embedding similarity to the reference \citep{nicolson2024longitudinal,rui2025improving}, we define our reward as
\begin{equation}
\begin{aligned}
r(y)\;=\;&0.20\,\mathrm{acc}(\hat{y},y^\ast)
\;+\;0.10\,J_{\mathrm{len}}(\hat{y},y^\ast)
\;+\;0.05\,J_{\mathrm{str}}(\hat{y},y^\ast)
\;+\;0.25\,\tfrac{Q_{\mathrm{ans}}(\hat{y},y^\ast)}{5}\\
&+\;0.15\,\tfrac{Q_{\mathrm{rsn}}(y,y^\ast)}{5}
\;+\;0.20\,\mathbf{1}[\mathrm{boxed}(y)]
\;+\;0.05\,\mathrm{sim}(\hat{y},y^\ast),
\end{aligned}
\label{eq:reward}
\end{equation}
where $\mathrm{acc}$ is normalized exact match, $J_{\mathrm{len}}$ and $J_{\mathrm{str}}$
judge whether the answer names the same disease and matches the reference's diagnostic
specificity, $Q_{\mathrm{ans}},Q_{\mathrm{rsn}}\in[1,5]$ grade the answer and the reasoning and
give partial credit, and $\mathrm{sim}\in[0,1]$ is embedding similarity to the reference. The weights stay
fixed throughout training.

%% file: figures/task_example.tex
\begin{figure*}[t]
\centering
\includegraphics[width=\textwidth]{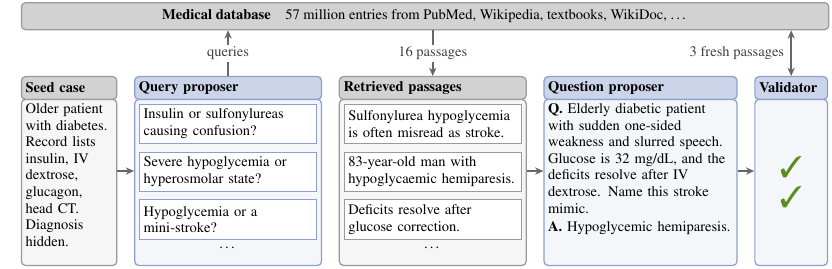}
\caption{\textbf{Example generated task workflow.} Blue boxes are agents, frozen
copies of the base model, and gray boxes are data. The example comes from the generation archive of
the 27B run, about a third of the way through RL training. Text is summarized, and the seed
admission is described only at a high level.}
\label{fig:task-example}
\end{figure*}

%% file: figures/base_vs_self.tex
\begin{figure*}[t]
\centering
\includegraphics[width=\textwidth]{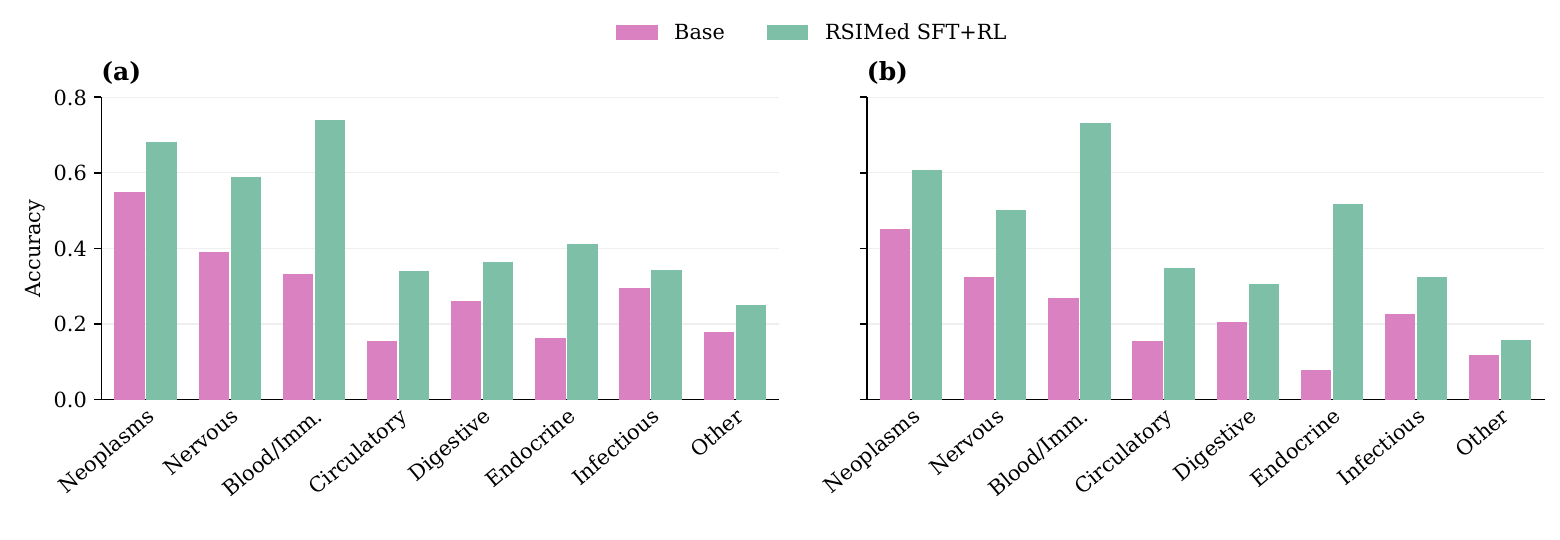}
\vspace{-2em}
\caption{\textbf{(a) Qwen3.6-27B diagnostic accuracy.} RSIMed reaches 0.516 overall, up from 0.334 for the untrained base model. \textbf{(b) Qwen3.5-9B diagnostic accuracy.} RSIMed reaches 0.488 overall, up from 0.261. Both models improve over their bases in every ICD chapter. The largest gains are in blood and immune disorders, which rise from 0.332 to 0.740 at 27B and from 0.269 to 0.735 at 9B.}
\label{fig:base-vs-self}
\end{figure*}

%% file: sections/experiments.tex
\section{Experiments}
\label{sec:experiments}

We design experiments to answer three research questions: \textbf{RQ1.} Does RSIMed keep improving in-domain diagnosis as training compute grows? \textbf{RQ2.} Does the improvement generalize to out-of-distribution clinical tasks? \textbf{RQ3.} How do the source and quality of generated tasks shape what the solver learns?

\input{tables/summary}

\subsection{Experimental Setup}
\label{sec:setup}

\textbf{Datasets and Training Setups.} For RQ1, we use MIMIC-Rare, a disease-diagnosis dataset sampled from MIMIC-IV \citep{johnson2023mimic} by matching International Classification of Diseases (ICD) codes to the rare-disease database Orphanet \citep{weinreich2008orphanet}. Its evaluation set contains 2,452 cases across eight ICD chapters. The training split supplies the labeled cases and the generation seeds, and evaluation admissions are never used for generation. As a control, we also train the supervised warm start with RL directly on the labeled training cases. For RQ2, we evaluate the models trained on rare-disease diagnosis, with no further training, on HealthBench Professional \citep{hicks2026healthbench}, 525 physician-written consultation, research, and writing tasks that supply no training data. For RQ3, we replace the self-generated task stream with a static pool of tasks generated once by the same proposer, the fixed-corpus setting of MedVLSynther \citep{huang2026synthesizing}, and with tasks from four external teachers, Kimi-K2.6, DeepSeek-V4-Pro, GPT-5.5, and Qwen3.5-397B. We report the last checkpoint of every run that does not collapse. Only baseline runs collapse, and for these we report the last checkpoint before the collapse so that the comparison stays meaningful.

\textbf{Metrics.} For MIMIC-Rare, we report diagnostic accuracy over the $N$ evaluation cases, $\mathrm{Acc} = \tfrac{1}{N}\sum_{i=1}^{N} J(\hat{y}_i, y_i^\ast)$, where $J\in\{0,1\}$ is a fixed LLM judge's verdict on whether the predicted disease $\hat{y}_i$ matches the reference $y_i^\ast$, allowing equivalent phrasing and code granularity. For HealthBench Professional, each task has a physician-written rubric of criteria $j$ with points $p_j$, positive for required content and negative for errors. A response $y$ meeting the criteria $M(y)$ receives $u(y) = \sum_{j\in M(y)} p_j \big/ \sum_{j\mid p_j>0} p_j$ and $s(y) = u(y) - \lambda\,(\ell(y)-\ell_0)/500$, where $\ell(y)$ is the answer length in characters, $\ell_0=2{,}000$, and $\lambda=0.0147$. We report the task means of $s(y)$ as the official adjusted score and of $u(y)$ as the raw score. GPT-5.5-Chat \citep{singh2025openai} grades all datasets.

\textbf{Models and baselines.} We use Qwen3.6-27B and Qwen3.5-9B \citep{yang2025qwen3} as solvers. Frozen-backbone baselines include direct inference, MedRAG \citep{zhao2025medrag}, RAG-Fusion \citep{rackauckas2024rag}, i-MedRAG \citep{xiong2024improving}, five-sample self-consistency, self-refinement \citep{madaan2023self}, and chain-of-verification \citep{dhuliawala2024chain}. We also compare with off-the-shelf GPT-5.6 Luna, Terra, and Sol, which receive no task-specific training, each at five reasoning efforts. Inference costs use API prices for GPT and the serving rates of Table~\ref{tab:pricing} for RSIMed. Due to compute constraints, we perform no hyperparameter tuning, and all baselines use the same hyperparameters. Details are included in Appendices~\ref{app:inference-baselines} and~\ref{app:training}.

\input{figures/hb_examples}

\subsection{RQ1. In-Domain Scaling with Training Compute}
\label{sec:rq1}\label{sec:results-short}

\textbf{Accuracy grows linearly with training compute.} Figure~\ref{fig:teaser}(b) shows approximately linear improvement for RSIMed-27B over the observed compute range, from 0.421 to 0.516, about 3.9 points per $10^{20}$ FLOPs. The 9B model keeps improving after $5\times10^{19}$ FLOPs, from about 0.46 to 0.488 at its last checkpoint after $1.4\times10^{20}$ FLOPs, with each step costing about a third of a 27B step.

\textbf{RL on a fixed task set saturates and then collapses, while RSIMed keeps improving.} From the warm start of RSIMed-27B, RL on the labeled training cases peaks at 0.463 and then collapses to 0.133, and RL on the static pool peaks at 0.441 and declines to 0.359, as Figure~\ref{fig:teaser}(b) shows. Both sets have limited data diversity, so once the solver has learned what they hold, further updates add no new knowledge. RSIMed instead keeps writing new tasks from retrieved evidence.

\textbf{The RL stage lifts RSIMed above off-the-shelf frontier GPT models at a fraction of their cost.} Figure~\ref{fig:teaser}(a) compares RSIMed with GPT-5.6 on the cost of answering all 2,452 evaluation cases. RSIMed-27B reaches 0.516 for \$0.57 and RSIMed-9B 0.488 for \$0.17. The most accurate GPT-5.6 setting, Luna at its highest reasoning effort, reaches 0.399 for \$21.92, 38 times the cost of RSIMed-27B, and the most expensive, Sol at its highest effort, reaches 0.383 for \$254.94. Most of this margin comes from RL, which adds 19.9 points to the 9B warm start at 0.288, 11 points below the best GPT-5.6 setting, and 9.5 points to the 27B warm start at 0.421. GPT-5.6 gains only 4.8 to 7.3 points from no reasoning to the highest effort, while its cost grows 7 to 18 times. In other words, thinking longer does not close the gap that RL on the task opens. Appendix~\ref{sec:cost} gives the token counts.

\textbf{RSIMed improves accuracy in all eight disease groups, while inference-time methods add at most 1.4 points.} Figure~\ref{fig:base-vs-self} shows that both RSIMed models improve over their bases in all eight disease groups, and Figure~\ref{fig:delta} shows gains in every group over static-pool training as well, for example from 0.565 to 0.740 on blood and immune disorders. The best frozen-backbone inference baseline, five-sample self-consistency, reaches 0.272 at 9B and 0.346 at 27B, against direct-inference scores of 0.262 and 0.332. Table~\ref{tab:main} gives every baseline by chapter.

\input{figures/task_source}

\subsection{RQ2. Out-of-Distribution Generalization to Professional Clinical Tasks}
\label{sec:rq2}

\textbf{Training only on diagnosis raises HealthBench Professional scores, most on difficult prompts and on consultation and writing.} RSIMed-RL-27B improves its backbone from 0.291 to 0.303 adjusted score and from 0.363 to 0.376 raw score, as shown in Figure~\ref{fig:summary}(b) and (c). Prompt by prompt, the raw score rises by 0.024 on the 269 prompts labeled difficult and by 0.002 on the 256 typical ones. Table~\ref{tab:hb-slices} shows that the adjusted gain is larger for consultation and writing, which rise from 0.261 to 0.277 and from 0.158 to 0.175, and red-teaming requests, whose framing invites an unsafe or incorrect answer, rise from 0.030 to 0.047.

\textbf{The transferred model outscores open medical models and the much larger DeepSeek-V4-Pro.} In Table~\ref{tab:main-hb}, RSIMed-RL-27B exceeds the other open base models, the open medical models MedGemma-1.5-4B, Lingshu-7B, and Lingshu-32B at 0.145 or below, and the much larger DeepSeek-V4-Pro at 0.282, while Kimi-K2.6 and Grok-4.3 remain ahead. The four teacher-trained RL variants in Table~\ref{tab:teachers} score between 0.289 and 0.308, so self-generated tasks transfer as well as teacher tasks.

\textbf{The trained model flags more clinical risks than base.} In Figure~\ref{fig:hb-examples}, the trained model makes a safer clinical decision than its base on two difficult prompts. It keeps a chest-pain patient with a rising troponin, a blood marker of heart injury, under observation where the base model discharges her, and it warns that a newly prescribed decongestant can raise uncontrolled blood pressure where the base model adds a fourth blood pressure medicine. In both, it notices a risk that the request does not raise.

\subsection{RQ3. How the Task Source Shapes Learning}
\label{sec:rq3}

\textbf{Surprisingly, on MIMIC-IV, self-improvement gives a larger performance boost than stronger teachers.} With RL only, self-generated tasks reach 0.350 MIMIC-IV accuracy against 0.309 to 0.344 for the four teacher sources, as shown in Figure~\ref{fig:task-source}(a), adding 1.6 points to the untrained student's 0.334 where the strongest teacher adds 1.0 point. Self-generated short-answer tasks also give the solver more to learn from. 36\% of them give a non-zero reward variance, against 22\% of GPT-5.5 tasks, and they are harder for the student, as shown in Figure~\ref{fig:task-source}(b), where 77\% of GPT-5.5 tasks sit at the all-correct end and the reward does not vary. Table~\ref{tab:teachers} and Appendix~\ref{app:teacher} give the full analysis.

\textbf{The proposer follows an increasingly fine-grained curriculum, asking more specific questions as the training proceeds.} Figure~\ref{fig:seed-evolution} follows the tasks for one training seed across four visits. We see the generated questions across training curriculum evolves from first-line drug class for alcohol withdrawal to the rescue agent when benzodiazepines fail, getting more fine grained and asking for more counterfactual cases each time the training progresses.

\input{figures/seed_evolution}

\textbf{An independent audit finds 84\% of accepted tasks valid, and the tasks cover diagnosis, mechanism, treatment, tests, and pharmacology.} An audit of 200 accepted and 200 rejected candidates from the 27B run, scored by GPT-5.5-Chat with databases, rates 84.1\% of the accepted tasks as valid and finds that the validator keeps 99.1\% of all valid tasks. The assessor is blinded to the validator's decision and checks each task and its proposed answer against freshly retrieved medical references. Figure~\ref{fig:task-source}(c) shows the clinical aspects the tasks cover, and although the proposer never sees the recorded diagnosis of its seed case, 29\% of the tasks concern that diagnosis and 25\% a clinically linked condition.

\textbf{SFT followed by RL gives the best in-domain performance, while RL alone gives the best out-of-distribution performance.} RL contributes more than the supervised warm start at both sizes. To separate the two stages, we evaluate the 27B model untrained, after the supervised warm start alone, after RL from the base weights, and after the full pipeline, on both benchmarks. The warm start raises MIMIC-IV accuracy from 0.334 to 0.421 at 27B and only from 0.261 to 0.288 at 9B, and RL after it reaches 0.516 and 0.488, as shown in Table~\ref{tab:stages}. On HealthBench Professional, however, RL from the base weights scores highest, at 0.303 adjusted, against 0.290 for the warm start alone and 0.280 for the full pipeline. Regraded case by case, the RL stage solves 329 evaluation cases that the supervised checkpoint missed and loses 98, a net gain of 231 cases across every ICD chapter. Since generation runs continuously, propose and solve rounds track training length, along which Figure~\ref{fig:teaser}(b) shows steady gains. Appendix~\ref{app:rl-stage} gives the full breakdown.

%% file: tables/summary.tex
\begin{figure*}[t]
\centering
\includegraphics[width=\textwidth]{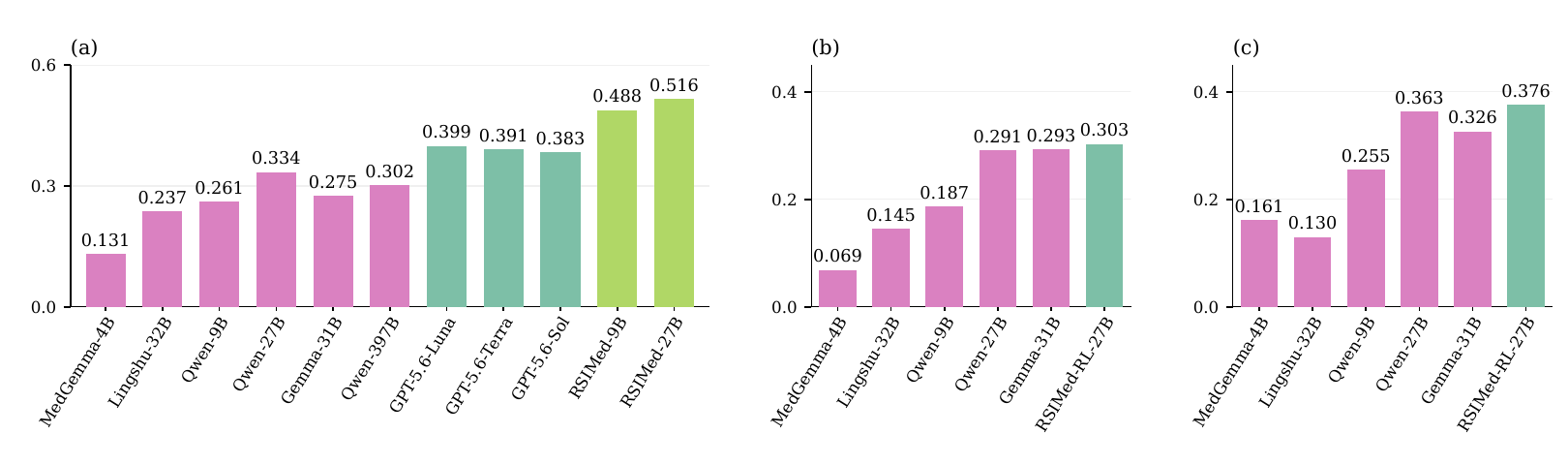}
\vspace{-2em}
\caption{\textbf{(a) MIMIC-IV model accuracy.} Trained on the task, RSIMed exceeds every displayed model, including off-the-shelf GPT-5.6.
\textbf{(b) HealthBench adjusted scores.} After rare-disease diagnosis training, RSIMed-RL-27B reaches $0.303$ on out-of-distribution HealthBench Professional tasks, exceeding the displayed baselines under GPT-5.5-Chat grading.
\textbf{(c) HealthBench raw scores.} The same RL-only variant reaches $0.376$, exceeding the displayed baselines.}
\label{fig:summary}
\end{figure*}

%% file: figures/hb_examples.tex
\begin{figure*}[t]
\centering
\includegraphics[width=\textwidth]{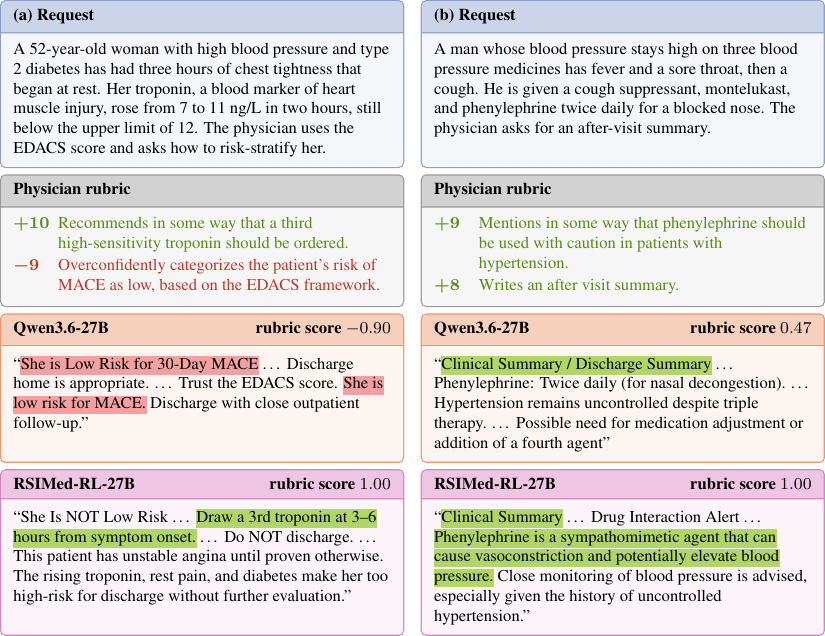}
\caption{\textbf{Two difficult HealthBench Professional prompts on which the trained model catches a
risk that its base model misses.} Requests are abridged, and responses are quoted with omissions
marked. Each physician rubric is listed in full, with positive criteria in green and negative
criteria in red. Green highlights mark response text that meets a positive criterion and red
highlights text that meets a negative one, following the verdicts of GPT-5.5-Chat, whose raw
rubric scores are shown for each response.}
\label{fig:hb-examples}
\end{figure*}

%% file: figures/task_source.tex
\begin{figure*}[t]
\centering
\includegraphics[width=\textwidth]{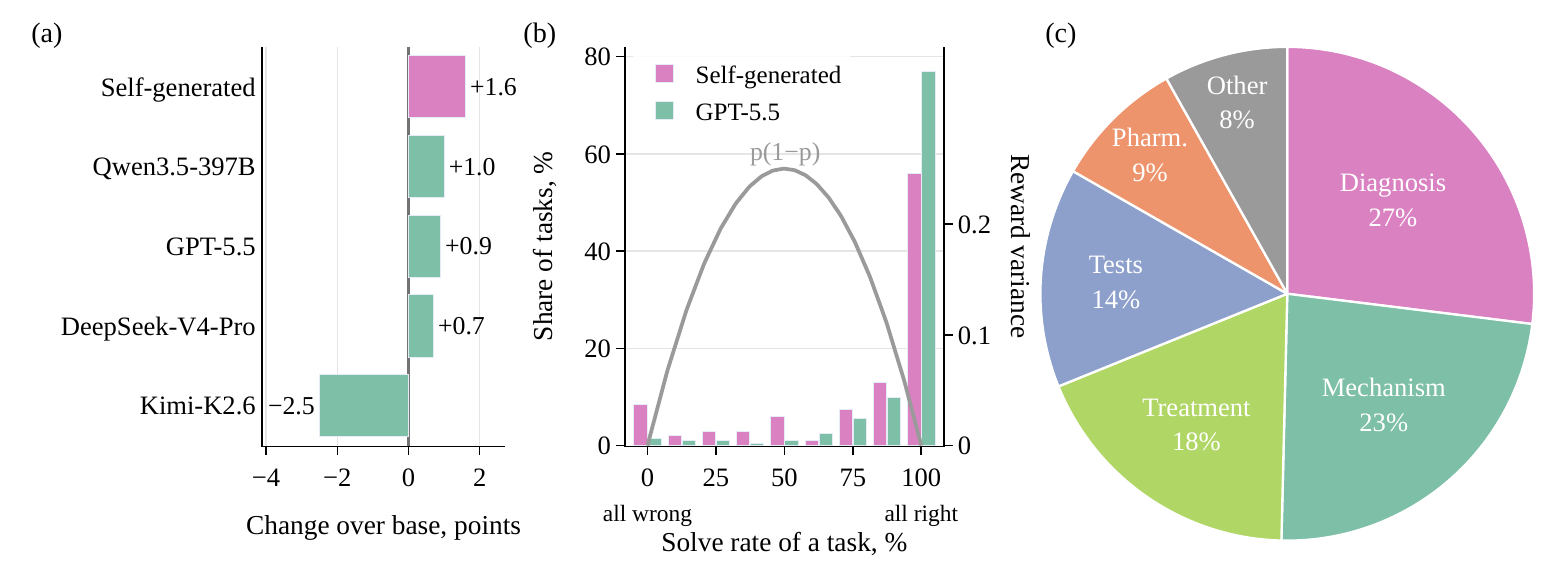}
\vspace{-1.5em}
\caption{\textbf{Self-generated tasks teach the solver more than tasks from stronger teachers and keep more tasks where the reward varies.} \textbf{(a)} Change in MIMIC-IV accuracy over the untrained base model after RL on tasks from each source, all trained from the base weights. \textbf{(b)} Share of short-answer tasks at each solve rate when the untrained student answers each task eight times without the retrieved passages, over 200 well-formed tasks per source, with the reward variance $p(1-p)$ on the right axis. Most GPT-5.5 tasks are solved in all eight responses, where the reward does not vary. \textbf{(c)} Clinical aspect of the accepted generated tasks, labeled by an independent classifier on 1,584 sampled tasks. Tests stands for tests and imaging and Pharm.\ for pharmacology, and Other groups prognosis, epidemiology, and unclassified tasks.}
\label{fig:task-source}
\end{figure*}

%% file: figures/seed_evolution.tex
\begin{figure*}[t]
\centering
\includegraphics[width=\textwidth]{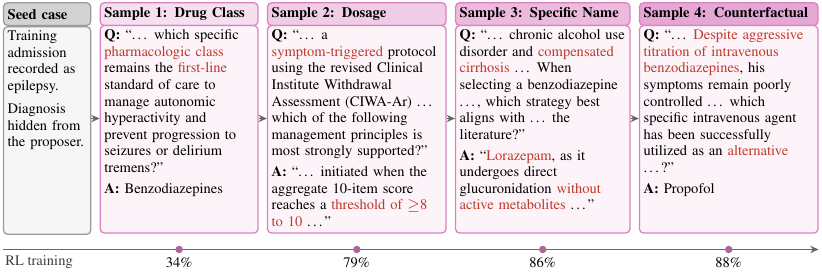}
\vspace{-1.5em}
\caption{\textbf{Four tasks the proposer wrote for one training case, one sampled from each of its visits.}
Every visit to this case produced tasks on alcohol withdrawal, a common cause of seizures. From
visit to visit the question narrows from the first-line drug class to the score that triggers a
dose, the choice of drug in cirrhosis, and the rescue agent when benzodiazepines fail. Red text marks what
makes each question more specific than the one before, and the line below marks the share of RL
training completed at each visit. Quoted text is verbatim and abridged, with
omissions marked, and multiple-choice answers quote the keyed option.}
\label{fig:seed-evolution}
\vspace{-1em}
\end{figure*}

%% file: sections/conclusion.tex
\section{Conclusion}
RSIMed lets a proposer and a solver co-evolve on clinical tasks, the solver in its weights through reinforcement learning and a frozen proposer in context, writing every task from retrieved medical evidence. On rare-disease diagnosis, accuracy grows linearly with training compute where a static pool collapses, and both trained RSIMed models exceed every off-the-shelf GPT-5.6 configuration at a fraction of its cost. The model also improves on unseen professional clinical tasks, which supports co-evolving task generation as a way to scale medical models beyond limited clinical data.

%% file: sections/statements.tex
\subsection*{AI use statement}
In this work, we used generative AI tools to polish writing and implement our proposed methods. Language models are also part of the method and of the experiments. A frozen copy of the solver writes and validates the synthetic training tasks, GPT-5.5, DeepSeek-V4-Pro, Kimi-K2.6, and Qwen3.5-397B write tasks in the teacher comparison, and GPT-5.5-Chat judges answers on both benchmarks. This work contains no theorems or proofs, and translation and survey design do not apply to it. We have reviewed all AI-assisted work. Every reported number is generated by script from saved evaluation outputs. We take responsibility for the final content of this work, including text, claims, and artifacts produced with the aid of generative AI.

\subsection*{Ethics statement}
This work conducts no new research with human subjects. MIMIC-IV is a de-identified collection of hospital records that PhysioNet distributes to credentialed researchers, and we used it under its data use agreement. HealthBench Professional and the corpora of the medical database are publicly released. RSIMed-27B and RSIMed-9B are research models whose diagnoses have not been validated in clinical practice, and they should not inform patient care without such validation and physician oversight. The training and evaluation admissions come from a single academic medical center in the United States, so the models may reflect the patient population and practice patterns of that center. Appendix~\ref{app:availability} describes the data sources, and Appendix~\ref{app:limitations} discusses the limits of our evaluation.

\subsection*{Reproducibility statement}
Section~\ref{sec:method} defines each component of RSIMed together with its reward, and Section~\ref{sec:setup} gives the datasets, the rare-disease case selection, the metrics, the models, and the baselines. Appendix~\ref{app:pipeline} details the proposer, the medical database, and the validator. Appendix~\ref{app:training} lists the training hyperparameters and hardware, Appendix~\ref{app:judging} describes how answers are judged, and Appendix~\ref{app:inference-baselines} specifies the frozen-backbone baselines. Qwen3.6-27B and Qwen3.5-9B are openly released models, and credentialed researchers can obtain MIMIC-IV from PhysioNet. The code, prompts, and configurations are released in the public repository given in Appendix~\ref{app:availability}.

%% file: sections/appendix_related.tex
\section{Comparison with related methods}
\label{app:related}

Table~\ref{tab:related} compares RSIMed with the related methods discussed in Section~\ref{sec:related}.

\input{tables/related_features}

%% file: tables/related_features.tex
\begin{table}[!ht]
\centering
\caption{Related methods and the features that matter for self-improvement on clinical tasks. Model sizes gives the range of parameter counts of the trained models. Answer check is how answers are verified, and a database verifier re-checks each generated reference answer against evidence freshly retrieved from a medical database. Dynamic generation means new tasks are written throughout training, so the data grow with compute. Difficulty calibration means task difficulty is set by the solver's current accuracy. Evidence-grounded means each reference answer is written from retrieved documents. Free-text answers means open answers graded for meaning, with no execution, option letter, or string match. Self improve means the model generates its own training data, and no stronger model writes or checks it. Clinical records means tasks are seeded from or evaluated on real patient records. Beats frontier means the trained model outperforms a frontier proprietary model on its target task.}
\label{tab:related}
\footnotesize
\setlength{\tabcolsep}{4pt}
\newcommand{\hd}[2]{\begin{tabular}[b]{@{}c@{}}\textbf{#1}\\\textbf{#2}\end{tabular}}
\resizebox{\textwidth}{!}{%
\begin{tabular}{@{}llllccccccc@{}}
\toprule
\textbf{Method} & \textbf{Domain} & \hd{Model}{Sizes} & \hd{Answer}{Check} & \hd{Dynamic}{Generation} & \hd{Difficulty}{Calibration} & \hd{Evidence-}{Grounded} & \hd{Free-Text}{Answers} & \hd{Self}{Improve} & \hd{Clinical}{Records} & \hd{Beats}{Frontier}\\
\midrule
Absolute Zero \citep{zhao2026absolute} & Code, math & 3B--14B & execution & \cmark & \cmark & \xmark & \xmark & \cmark & \xmark & \xmark\\
R-Zero \citep{huang2026r} & Math, reasoning & 3B--8B & majority vote & \cmark & \cmark & \xmark & \xmark & \cmark & \xmark & \xmark\\
SPICE \citep{liu2025spice} & Math, general & 3B--8B & rule match & \cmark & \cmark & \cmark & \xmark & \cmark & \xmark & \xmark\\
WIST \citep{li-etal-2026-wist} & General, medical exams & 3B--14B & rule match & \cmark & \cmark & \cmark & \xmark & \cmark & \xmark & \xmark\\
MedVLSynther \citep{huang2026synthesizing} & Biomedical VQA & 3B--7B & model verifier & \xmark & \xmark & \cmark & \xmark & \xmark & \xmark & \xmark\\
HuatuoGPT-o1 \citep{chen2024huatuogpt} & Medical exams & 7B--70B & GPT-4o verifier & \xmark & \xmark & \xmark & \cmark & \xmark & \xmark & \xmark\\
SCALEMED \citep{yilmaz2025synthetic} & Dermatology & 11B & none & \xmark & \xmark & \xmark & \xmark & \xmark & \xmark & \cmark\\
EDCO \citep{pang2026edcodynamiccurriculumorchestration} & Medical, law & 3B--4B & answer key & \xmark & \cmark & \xmark & \xmark & \xmark & \xmark & \xmark\\
\midrule
RSIMed (ours) & Clinical diagnosis & 9B--27B & database verifier & \cmark & \cmark & \cmark & \cmark & \cmark & \cmark & \cmark\\
\bottomrule
\end{tabular}}
\end{table}

%% file: sections/appendix_pipeline.tex
\section{Proposer pipeline details}
\label{app:pipeline}

The proposer service handles query generation, retrieval, task generation, and validation.
The trainer consumes accepted items over HTTP while generation runs asynchronously. A pool
holds up to $200$ validated items, sampled in random order. Eight concurrent workers refill
it whenever it is below capacity. If generation falls behind, the service can reuse
previously accepted tasks from its history.

\textbf{Knowledge base.} The proposer retrieves from a vector index of medical
literature, textbook passages, and question-answer resources, embedded with
Qwen3-VL-Embedding-2B. The index holds eight source families, PubMed abstracts, Wikipedia, and medical textbooks from
the MedRAG corpus, WikiDoc, PMC-VQA \citep{zhang2023pmc}, MIRAGE, PubMedQA
\citep{jin2019pubmedqa}, and CLIMB \citep{dai2025climb}. Table~\ref{tab:kb-sources} lists their
sizes. Retrieved text retains its source label and is supplied to the
proposer as evidence for task generation. The diagnostic experiments use text questions.

\textbf{Query proposal and retrieval.} The proposer writes ten queries per seed, covering
mechanism, diagnostic criteria and workup, comparative effectiveness, adverse effects and
complications, prognosis and natural history, atypical presentation, differential diagnosis
against one or two named alternatives, imaging findings, epidemiology and risk factors, and
related conditions. Each query retrieves the sixteen nearest passages. These are concatenated
with source tags in the generation prompt. Query proposal uses temperature $0.8$;
question generation uses temperature $0.9$. Both calls allow up to $12{,}288$ output tokens.

\textbf{Training pool.} The pool combines labeled training cases with generated tasks.

\textbf{History conditioning.} Each accepted item is embedded into a history collection with
running solved and attempted counts. Before writing queries, the worker retrieves the ten most
similar previous questions and their solve rates, each shortened to its first 400 characters,
so this part of the prompt stays under 5,000 characters however long training runs. The prompt instructs the proposer to avoid
repetition and topics the solver answers reliably, and to target failed or untested topics.

\textbf{Validation.} The schema check rejects empty questions or answers, format mismatches,
and multiple-choice keys absent from the options. The model check retrieves the three nearest
passages for the finished question, including option text for multiple-choice items,
and evaluates it at temperature $0.2$ using the proposer model. It rejects incoherent questions and answers directly contradicted by the passages.
Inconclusive evidence leaves the candidate accepted. Rejections are logged with the
contradicting passage. Candidates are also retained when retrieval returns no passages
or the model check fails to return a verdict.

\textbf{Difficulty feedback.} As each rollout is scored, the trainer reports its exact-match
correctness. The service maintains a sliding accuracy mean over the latest $64$ reports and
per-question counts. Both enter the generation prompt with a target solve rate of one half,
and the side of the target on which the running accuracy lies sets the direction of the change.
Above the target, the proposer writes finer clinical distinctions and, for multiple-choice
questions, distractors closer to the correct answer. Below the target, it phrases its questions
more precisely while keeping the same reasoning step, so they become easier by removing ambiguity.

%% file: sections/appendix_generated_tasks.tex
\section{Analysis of generated tasks}
\label{app:generated-tasks}

This section describes what the proposer generated during the 27B SFT+RL run. The data are the
run's archive of accepted tasks and the correctness of every
rollout that the trainer reported back to the proposer. The reported correctness is the
exact-match signal that drives the difficulty feedback. Each report is assigned to the training
step whose logged completion time follows it. We analyze the $43{,}698$ well-formed generated
tasks that were served during training and received a full group of eight reports.

\textbf{The running accuracy that the proposer sees stays close to its target of one half.} The proposer sees the running accuracy over the
most recent rollouts, which pools both formats and the labeled cases. Averaged over each tenth
of the run, this running accuracy is $0.47$ in the first window and stays between $0.52$ and
$0.57$ afterwards, close to the target of one half, while the solver's exact-match accuracy on
labeled cases rises from $0.25$ to $0.36$ over the same period.

\textbf{Over half of the tasks concern the seed's recorded diagnosis or a linked condition, although the proposer never sees that diagnosis.} An independent classifier, GPT-5.5-Chat, labeled a
sample of $1{,}584$ tasks, drawn evenly across training, with the clinical aspect each task
tests and its relation to the recorded diagnosis of the seed case. The proposer never sees that
diagnosis. Table~\ref{tab:gen-topics} gives the shares.

\begin{table*}[htbp]
\centering\small
\caption{Clinical aspect of generated tasks and their relation to the seed case's recorded
diagnosis, from an independent classifier on $1{,}584$ sampled tasks.}
\label{tab:gen-topics}
\begin{tabular}{lr@{\hspace{2.5em}}lr}
\toprule
Aspect & Share & Relation to the seed's diagnosis & Share \\
\midrule
Diagnosis & 27.0\% & Same disease & 28.8\% \\
Mechanism & 23.4\% & Clinically linked condition & 25.4\% \\
Treatment & 18.5\% & Other detail of the case & 8.8\% \\
Tests and imaging & 14.4\% & No identifiable link & 37.0\% \\
Pharmacology & 8.6\% & & \\
Prognosis & 4.5\% & & \\
Epidemiology & 3.0\% & & \\
\bottomrule
\end{tabular}
\end{table*}

\textbf{On later visits to a case, the proposer narrows the questions of earlier visits.} The proposer returns to many training cases during the run, and
on a later visit it can take up a thread from an earlier one and narrow it.
Figure~\ref{fig:seed-evolution} shows one of the $129$ training cases that the proposer visited at
least three times over at least half of the run, chosen after reviewing every visit to all $129$.
All four visits to this case, a training admission recorded as epilepsy, produced tasks on alcohol
withdrawal. The first task asks for the first-line drug class. The later ones ask when to give a
dose under the CIWA-Ar score, which benzodiazepine to use in cirrhosis, and which agent to use when
benzodiazepines fail. Each question narrows the one before it to a more specific clinical decision.

\textbf{Medical textbooks are retrieved far more often than their share of the index.} Table~\ref{tab:kb-sources} lists the sources of the knowledge base with their
sizes, and Table~\ref{tab:kb-usage} shows how often each appears in the evidence of generated
tasks. The archive stores the
first $4{,}000$ characters of the evidence for each task, which hold its top-ranked passages.
The medical textbooks make up $0.2\%$
of the index but appear in the evidence of $28\%$ of the tasks.

\input{tables/kb_usage}

%% file: tables/kb_usage.tex
\begin{table}[htbp]
\centering\small
\caption{Use of each knowledge-base source in the evidence of generated tasks. Tasks citing is the
share of accepted generated tasks whose evidence includes the source, and top passage is the share
whose highest-ranked passage comes from it, over the 43{,}698 analyzed tasks of the 27B run.}
\label{tab:kb-usage}
\begin{tabular}{lrr}
\toprule
Source & Tasks citing & Top passage \\
\midrule
PubMed abstracts & 97.3\% & 61.9\% \\
Wikipedia & 30.5\% & 5.0\% \\
Medical textbooks & 28.2\% & 4.6\% \\
WikiDoc & 50.6\% & 25.0\% \\
PMC-VQA & 10.8\% & 3.5\% \\
MIRAGE & 0.2\% & $<$0.1\% \\
PubMedQA & $<$0.1\% & $<$0.1\% \\
CLIMB & $<$0.1\% & 0\% \\
\bottomrule
\end{tabular}
\end{table}

%% file: sections/appendix_training.tex
\section{Training configuration}
\label{app:training}

The GRPO runs use two to four B200 GPUs depending on the model size, FSDP2, and vLLM-backed asynchronous rollouts. Settings
include learning rate $2\times10^{-7}$, training batch size $64$, mini-batch size $32$,
eight rollouts per prompt, temperature $1.0$, maximum prompt length $8192$, and maximum
response length $4096$. The main 27B configuration uses the \texttt{kl\_cov} policy loss,
which penalizes changes in token log probability for tokens with the highest covariance
between log probability and advantage. Its selection ratio is $0.001$ and penalty coefficient
is $1.0$. An overlong-response penalty applies near the response limit. The full pipeline initializes the actor from the merged
supervised checkpoint; RL-only comparisons initialize from base weights. The proposer and
judge use separately served frozen models. Validation uses greedy decoding at a fixed cadence.
We report the last checkpoint of every run that does not collapse. Only baseline runs collapse, and for these we report the last checkpoint before the collapse so that the comparison stays meaningful.

The scaling runs initialize from the supervised checkpoint.

\paragraph{Warm-start trace generation.} For a trace $\tau$ with extracted answer
$\hat{y}$ and reference $y^\ast$, we accept the trace when $A(\tau)=1$, where
\begin{equation}
A(\tau)\;=\;
\underbrace{\mathbf{1}[\mathrm{fmt}(\tau)]}_{\text{think-then-box format}}\cdot
\underbrace{\mathbf{1}[\mathrm{icd}(\hat{y},y^\ast)]}_{\text{ICD-aware match}}\cdot
\underbrace{\mathbf{1}[\tau\notin\mathcal{L}]}_{\text{no leak}}\cdot
\underbrace{\mathbf{1}[\ell_{\min}\le|\tau|\le\ell_{\max}]}_{\text{length}}\cdot
\underbrace{\mathbf{1}[\max_{\tau'\in\mathcal{A}}\mathrm{sim}(\tau,\tau')<\delta]}_{\text{not a near-duplicate}}
\label{eq:sftgate}
\end{equation}
Here $\mathrm{fmt}$ checks the reasoning and boxed-answer format. When both answers contain
ICD codes, $\mathrm{icd}$ accepts equal codes or parent--child prefix matches; otherwise it
uses case-insensitive equality or substring matching. The pattern set $\mathcal{L}$ detects
explicit references to the supplied diagnosis, such as ``the provided diagnosis'' or
``working backward.'' The set $\mathcal{A}$ contains the case's previously accepted traces.
We sample traces at temperature $0.8$ in rounds of four candidates, with at most three
rounds per case, until three traces pass the filters. Each trace must contain at least
$200$ characters of reasoning and between $400$ and $1400$ estimated tokens. We estimate
length at $3.6$ characters per token. Thirteen case-insensitive patterns detect references
to the supplied diagnosis. We also strip parenthetical or labeled ICD codes from the
reasoning and reject traces with remaining codes. To remove near-duplicates, we compare
the first $1200$ characters of normalized reasoning with previously accepted traces using
a sequence-match threshold of $0.7$.

\paragraph{Reward grading.} Judge calls use deterministic decoding with reasoning disabled.
We penalize responses that enter the final $512$ tokens of the $4096$-token response budget.
The composite reward weights remain fixed.

\paragraph{Compute accounting.} We estimate the FLOPs of a forward pass over a sequence of $T$ tokens as
\begin{equation}
F[T] \;=\; \underbrace{2 N_{\mathrm{act}} T}_{\text{parameters}} \;+\;
           \underbrace{2 L d\, T^{2}}_{\text{attention}},
\label{eq:flops}
\end{equation}
where $N_{\mathrm{act}}$ is the number of active parameters, $L$ the layer count, and $d$ the
model dimension. The parameter term includes prefill and decoding with cached keys and values.
The attention term sums $4Ld\,i$ FLOPs over token positions $i$. It contributes under $2\%$
at our rollout lengths, $10\%$ at $33$k tokens, and $29\%$ at $124$k tokens.

For every training curve in Figure~\ref{fig:teaser}(b), the per-step cost sums rollout
generation, the policy update, task proposal, and reward judging. Rollout generation uses
$F$ over the step's rollouts. The policy update costs $6NT$ for the forward-backward
pass plus $2NT$ for the log-probability pass. We use $Q=64$
proposed tasks and $R=512$ rollouts per step. $N$, $L$, and $d$ are those of the model being
trained, since proposer, solver, and judge share one base model. We use $N=2.78\times10^{10}$,
$L=64$, $d=5120$ for Qwen3.6-27B and $N=9.65\times10^{9}$, $L=32$, $d=4096$ for Qwen3.5-9B,
so a 9B step costs roughly a third of a 27B step and the two curves span different compute
ranges for the same number of steps. The
static-pool run incurs its proposal cost once, before training, for the $15{,}000$ tasks in its frozen
pool. This costs $2N\cdot15{,}000\cdot\tau_{\mathrm{prop}}$, about $8\times10^{18}$ FLOPs or approximately twenty
training steps, with no further proposal cost during training. The training-set run has no proposal cost. Token counts come from the training log except
for the proposer and judge, which run on a separate inference server that does not log them;
those two use $\tau_{\mathrm{prop}}=10{,}000$ tokens per accepted task, covering retrieved passages,
instructions, the generated task, and rejected drafts, and $\tau_{\mathrm{judge}}=1{,}600$
tokens per rollout. Together they account for about a quarter of estimated compute. The run spans
$1.7\times10^{18}$ to $2.7\times10^{20}$ FLOPs; the supervised warm start adds under one
percent and is omitted.

\paragraph{GPU hours.} We convert FLOPs to GPU hours with the measured wall-clock time of the
training steps, which excludes validation passes, as the FLOP count does. RSIMed-27B trains on
four B200 GPUs for 222 hours, or 887 GPU hours over $2.75\times10^{20}$ FLOPs, which gives 323 GPU
hours per $10^{20}$ FLOPs. RSIMed-9B trains for 153.5 hours on two B200 GPUs and then 42.3 hours on four, or 476 GPU hours over
$1.45\times10^{20}$ FLOPs, which gives 329 GPU hours per $10^{20}$ FLOPs. Both models therefore
need about 325 B200 GPU hours per $10^{20}$ FLOPs. These hours count the GPUs of the training node;
the proposer and judge run on a separate inference server.

%% file: sections/appendix_baselines.tex
\section{Frozen-Backbone Inference Baselines}
\label{app:inference-baselines}

All methods use the same $2452$ MIMIC-IV cases, clinical text, and available images
as the direct evaluation. Model parameters remain fixed. The original clinical
text cap is $9000$ characters, image resolution is capped at $65{,}536$ pixels per
image, and the final answer budget is $8192$ tokens. Qwen's thinking mode is
disabled, matching the reference evaluation. The final answer extractor and
judge are fixed across methods. Ground-truth diagnoses are used only for grading;
they are not available to retrieval, planning, critique, or answer selection.

\paragraph{Shared retrieval.}
Medical RAG, RAG-Fusion, and i-MedRAG run on a common corpus and backbone. Retrieval
uses the complete medical database of Table~\ref{tab:kb-sources}, with the same embedding
model and cosine search as the proposer. We retrieve $32$ candidates, reject duplicate passages
and fragments shorter than $80$ characters, retain up to eight passages, and cap
each retained passage at $1200$ characters. If no usable passage survives, the
candidate depth increases to $128$ and then $512$ while keeping the final context
budget fixed. If the search still returns no usable evidence,
the model is told that references are unavailable.

\paragraph{Retrieval methods.}
Single-pass medical RAG retrieves using the case text and supplies the resulting
passages to the solver. RAG-Fusion adds three model-generated search queries,
combines their rankings with the original query using reciprocal-rank fusion
with constant $60$, and retains eight passages for the final answer. i-MedRAG
performs three rounds of two follow-up questions. Each follow-up is answered with
retrieval, using at most $1024$ answer tokens. The accumulated question-answer
pairs guide the next round and the final diagnosis. The same frozen backbone
generates the questions, answers the follow-ups, and makes the final prediction.

\paragraph{Other inference methods.}
Self-consistency samples five independent answers at temperature $0.7$ and selects
the most frequent extracted diagnosis after lowercasing and punctuation
normalization; ties select the earliest sample. Self-refinement produces a draft, a self-critique against the
case evidence, and a revised diagnosis. Chain-of-verification produces a draft
and three factual verification questions, answers each independently without
showing the draft, and revises the diagnosis using those answers. Critiques and
verification answers are capped at $1024$ tokens. Other than self-consistency,
generation uses temperature zero.

%% file: sections/appendix_judging.tex
\section{Judging}
\label{app:judging}

MIMIC-IV comparisons use stored generations regraded with a fixed prompt and GPT-5.5-Chat.
Regrading covers all $197$ validation passes of the continuous 27B run, all $32$ passes of the
static-pool control, and every fourth pass of the 9B run through its last checkpoint. Each pass contains $2452$ cases.
The training-set curve in Figure~\ref{fig:teaser}(b) and its row in Table~\ref{tab:main} use the
GPT-5.5-Chat grades recorded during training.

HealthBench Professional headline results and the slice analysis in
Table~\ref{tab:hb-slices} use GPT-5.5-Chat with the official grading protocol. Slice scores are recomputed from all $525$ saved per-example grades
and verified against each run's official overall score. Length adjustment is applied
per example before averaging; each group mean is clipped to $[0,1]$.

\paragraph{Task-validator audit.}
We uniformly sample $200$ historically accepted and $200$ rejected candidates from
the full-pipeline generation archive. Sampling uses seed $42$. An independent GPT-5.5-Chat assessor,
blinded to the historical acceptance decision, labels each task valid, invalid, or
uncertain using its stem, proposed answer, and freshly retrieved medical references.
Of the $200$ accepted rows, $147$ are generated tasks and $53$ are labeled training cases, which
the pool serves without validation. Excluding uncertain assessments, $84.1\%$ of the accepted
generated tasks are valid, and the validator keeps $99.1\%$ of all valid tasks, weighting the
two samples by the sizes of the accepted and rejected populations.

\paragraph{Pricing.} Figure~\ref{fig:teaser}(a) uses measured token counts and the rates in
Table~\ref{tab:pricing}.

\input{tables/pricing}

%% file: tables/pricing.tex
\begin{table*}[htbp]
\caption{Rates used to estimate the inference costs in Figure~\ref{fig:teaser}(a), in US dollars per million
tokens. The 9B rate scales the 27B rate by
$9.65\text{B}/27.78\text{B}=0.35$. Frontier-model rates use API list prices, with billed
reasoning tokens included in output counts.}
\label{tab:pricing}
\begin{center}
\begin{tabular}{lcc}
\toprule
Model & Input & Output \\
\midrule
\method{}-27B & 0.05 & 0.30 \\
\method{}-9B  & 0.017 & 0.104 \\
GPT-5.6-luna         & 0.20 & 1.20 \\
GPT-5.6-terra        & 2.00 & 12.00 \\
GPT-5.6-sol          & 5.00 & 30.00 \\
\bottomrule
\end{tabular}
\end{center}
\end{table*}

%% file: sections/appendix_results.tex
\section{Additional clinical results}
\label{app:results}\label{app:variant}\label{app:teachers}
Diagnostic results use the full RSIMed-27B pipeline unless a different training variant is named. HealthBench results for our method use RSIMed-RL-27B, graded by GPT-5.5-Chat, because RL from the base weights gives the best HealthBench Professional score of the training variants in Table~\ref{tab:stages}, while the full pipeline gives the best MIMIC-IV accuracy.

The full 27B model exceeds its base in every ICD chapter, as Table~\ref{tab:main} shows. Table~\ref{tab:hb-slices} reports the HealthBench Professional breakdown, where RSIMed-RL-27B improves over its base on all three use cases and all three slices. The red-teaming score is 0.047 versus 0.030.
\input{figures/controls}

\input{tables/healthbench_slices}
\input{tables/teacher_comparison}

\input{tables/kb_ablation}

%% file: figures/controls.tex
\begin{figure}[htbp]
\centering
\includegraphics[width=\textwidth]{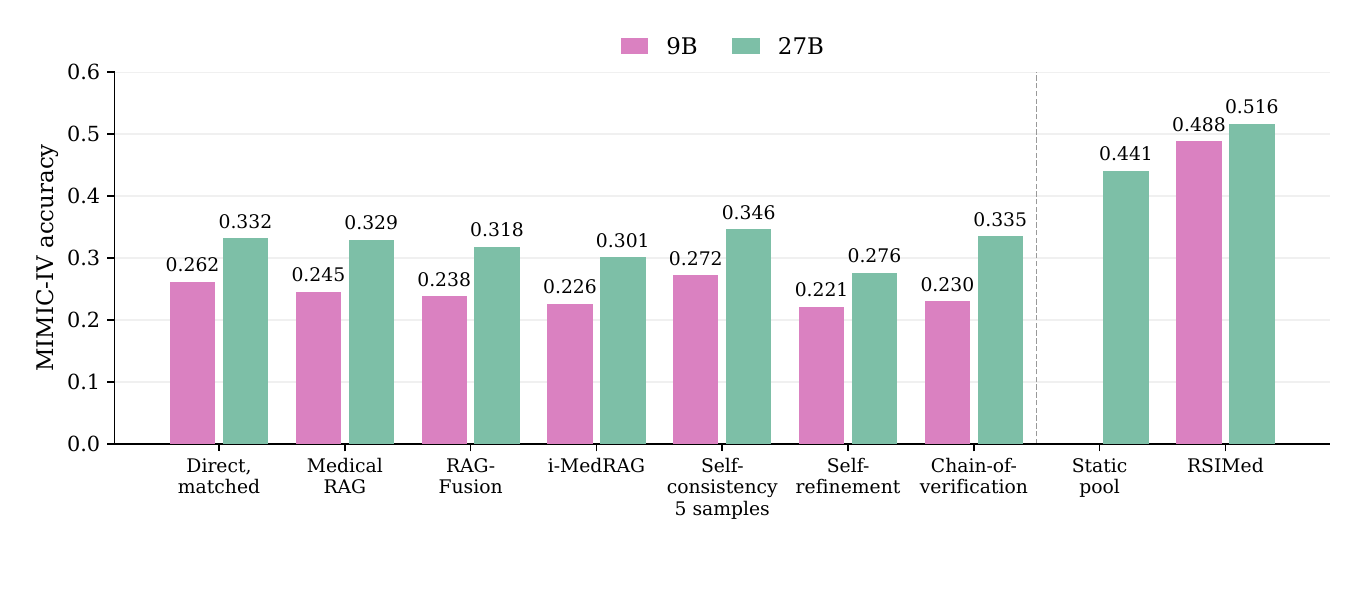}
\caption{\textbf{MIMIC-IV inference and training controls.} RSIMed attains the highest accuracy on both backbones, compared with direct inference and six alternative inference methods on the same backbones. The dashed line separates the frozen-backbone methods from SFT+RL training.}
\label{fig:controls}
\end{figure}

%% file: tables/healthbench_slices.tex
\begin{table*}[htbp]
\caption{HealthBench Professional by use case and dataset slice, length-adjusted,
graded by GPT-5.5-Chat, for the variants reported in Table~\ref{tab:main-hb}.
Group sizes are consult $236$, writing $142$, research $147$;
good-faith typical $256$, good-faith difficult $78$, red-teaming difficult $191$.}
\label{tab:hb-slices}
\centering\small
\setlength{\tabcolsep}{4pt}
\resizebox{\textwidth}{!}{\begin{tabular}{lccccccc}
\toprule
Model & Overall & Consult & Writing & Research & GF typ. & GF diff. & RT diff. \\
\midrule
Qwen3.6-27B base & 0.291 & 0.261 & 0.158 & 0.467 & 0.521 & 0.174 & 0.030 \\
RSIMed-RL-27B & 0.303 & 0.277 & 0.175 & 0.469 & 0.522 & 0.212 & 0.047 \\
\bottomrule
\end{tabular}}
\end{table*}

%% file: tables/teacher_comparison.tex
\begin{table*}[htbp]
\caption{Performance by training-data source. The source comparison uses RL without a supervised
warm start. Both benchmarks are graded by GPT-5.5-Chat. ``w/ $X$'' denotes training
with teacher $X$.}
\label{tab:teachers}
\begin{center}
\small
\begin{tabular}{lccc}
\toprule
& MIMIC-IV & \multicolumn{2}{c}{HealthBench Prof.} \\
\cmidrule(lr){2-2}\cmidrule(lr){3-4}
Training data source & Overall & Len-adj. & Raw \\
\midrule
Qwen3.6-27B, untrained   & 0.334 & 0.291 & 0.363 \\
\midrule
w/ Kimi-K2.6             & 0.309 & 0.298 & 0.366 \\
w/ DeepSeek-V4-Pro       & 0.341 & 0.289 & 0.362 \\
w/ GPT-5.5               & 0.343 & 0.294 & 0.366 \\
RSIMed-RL-27B & - & \textbf{0.303} & \textbf{0.376} \\
\midrule
\textbf{\method{}-27B, full pipeline (ours)} & \textbf{0.516} & -- & -- \\
\bottomrule
\end{tabular}
\end{center}
\end{table*}

%% file: sections/appendix_teacher_analysis.tex
\section{Teacher analysis with protocol, full measurements, and examples}
\label{app:teacher}

We analyze the task pools consumed by the five training runs using the untrained student and
three additional solvers. We report the protocol, measurements, and example tasks below. A teacher here supplies generated training tasks, while the student is the model trained on those tasks. Teacher quality therefore concerns what its examples teach the student, in addition to the teacher's own ability to answer questions.

\subsection{Protocol}
\label{app:teacher-protocol}

\paragraph{Task pools.} For each source, we retain tasks served during training with a stem of at
least eight words and a non-empty key. We exclude malformed items, including JSON placeholder
stems from the student's proposer. We sample $200$
multiple-choice and $200$ short-answer tasks uniformly from each filtered pool.

\paragraph{Learning signal.} Each task was answered eight times by the untrained student with the
training-time sampling settings, using temperature 1.0, top-$p$ 1.0, repetition penalty 1.1,
a 4096-token budget, and the same system prompts as the trainer. Multiple-choice answers were
scored by letter match; short answers were scored against the proposer's key by the reference
grader of Table~\ref{tab:main}, GPT-5.5-Chat with a lenient rubric. We report the fraction of groups with unequal correctness rewards and the mean within-group
variance $p(1-p)$, where $p$ is the observed correct-answer fraction. Equal-reward groups have
zero group-relative advantage. This correctness-based audit estimates learning opportunities;
the training reward also includes the terms in Equation~\ref{eq:reward}.
\paragraph{Solvers.} The same 800 tasks written by the student and by GPT-5.5 were also answered
eight times each by Gemma-4-31B, by Qwen3.5-9B with a 12k-token budget, and by GPT-5.5 itself with eight samples per call and low
reasoning effort. A further 400 multiple-choice tasks per proposer were answered by the
student, Gemma-4-31B, and GPT-5.5, which brings the multiple-choice test to 600 tasks per proposer.
``Sound'' tasks are those GPT-5.5 answers correctly in all eight samples.

\paragraph{Passage dependence.} Each short-answer task was answered again by the student with
the passages the proposer had retrieved placed above the question. The accuracy gain measures how much the retrieved context helps the student answer the task.

\paragraph{Stem leakage.} A stem leaks its key when a content word of the key appears in the stem, provided that the word has five letters or
more and is not a generic clinical word; for multiple choice the key is the
text of the correct option.

\paragraph{Redundancy and coverage.} Tasks were embedded with the project's Qwen3-VL-Embedding-2B retrieval embedder. On a random 900-task subset per pool we report the share of tasks whose
nearest neighbour exceeds cosine 0.9, which we count as near-duplicates, the number of greedy clusters at cosine
0.8, and the participation ratio of the embedding covariance; coverage is the share of the 2452
evaluation admissions with some task above cosine 0.7.

\begin{table*}[htbp]
\caption{Share of eight-rollout groups with unequal rewards, expressed as a percentage, and accuracy, when the tasks
written by the student and by GPT-5.5 are answered by four different solvers. ``Sound'' denotes tasks GPT-5.5 answers correctly in all eight samples. GPT-5.5 therefore
has zero reward variance on this subset.}
\label{tab:solver-matrix}
\begin{center}
\small
\setlength{\tabcolsep}{5.5pt}
\begin{tabular}{llcccccc}
\toprule
& & \multicolumn{3}{c}{Tasks by GPT-5.5} & \multicolumn{3}{c}{Tasks by the student} \\
\cmidrule(lr){3-5}\cmidrule(lr){6-8}
Solver & Format & Acc. & Signal & Sound signal & Acc. & Signal & Sound signal \\
\midrule
GPT-5.5 & MCQ & 1.000 & 0 & 0 & 0.978 & 2 & 0 \\
GPT-5.5 & Short & 0.956 & 1 & 0 & 0.849 & 13 & 0 \\
Gemma-4-31B & MCQ & 0.991 & 1 & 2 & 0.976 & 2 & 2 \\
Gemma-4-31B & Short & 0.911 & 16 & 17 & 0.786 & 22 & 19 \\
Qwen3.6-27B student & MCQ & 0.985 & 11 & 11 & 0.941 & 22 & 21 \\
Qwen3.6-27B student & Short & 0.925 & 22 & 22 & 0.787 & 36 & 31 \\
Qwen3.5-9B & MCQ & 0.978 & 8 & 5 & 0.929 & 17 & 14 \\
Qwen3.5-9B & Short & 0.851 & 22 & 20 & 0.706 & 40 & 37 \\
\bottomrule
\end{tabular}
\end{center}
\end{table*}
On the expanded multiple-choice comparison with $600$ tasks per proposer, of which $300$
written by GPT-5.5 and $293$ written by the student are sound,
accuracy on the sound subset is $0.977$ for the student on GPT-5.5's tasks and $0.959$ on
its own. Gemma-4-31B scores $0.995$ and $0.980$, respectively. Self-generated tasks
are harder for both solvers.

\subsection{Examples}
\label{app:teacher-examples}

Tasks and answers below retain their clinical content, with punctuation and abbreviation
presentation normalized for readability. The eight student answers are the extracted boxed
answers of the eight rollouts, in sampling order. The examples illustrate differences in generated task difficulty and answer variation. Correctness counts refer to the specified reference grader.

\paragraph{Self, Qwen3.6-27B.}
A multiple-choice task the student answers correctly in all eight rollouts. The reference key is B.
\begin{quote}\small A patient presents with severe pancytopenia and marked reticulocytopenia. To differentiate acquired aplastic anemia from myelodysplastic syndrome, bone marrow examination and cytogenetic analysis are performed. Which combination of findings most specifically supports a diagnosis of aplastic anemia?
\begin{itemize}\itemsep0pt
\item[A.] Hypercellular marrow with multilineage dysplasia and recurrent chromosomal translocations
\item[B.] Hypocellular marrow largely replaced by adipose tissue with normal karyotype
\item[C.] Normocellular marrow with increased reticulin fibrosis and ring sideroblasts
\item[D.] Hypercellular marrow with \textgreater{}20\% circulating blasts and cytogenetic abnormalities
\end{itemize}\end{quote}
A short-answer task with mixed student correctness has 6 of 8 answers accepted by the reference grader.
\begin{quote}\small In Crohn's disease management, diffusion-weighted magnetic resonance enterography is often used to differentiate active inflammatory strictures from fibrotic strictures to guide medical versus surgical therapy. Which quantitative MRI parameter is primarily measured for this distinction, and how does its value characteristically change during active inflammation?\\[2pt]
\emph{Reference answer.} Apparent diffusion coefficient decreases\\
\emph{Student answers.} Decreased apparent diffusion coefficient; Decreased apparent diffusion coefficient; Apparent diffusion coefficient; increases; acute pancreatitis; Decreased apparent diffusion coefficient; Apparent diffusion coefficient, which decreases; Apparent diffusion coefficient decreases; Apparent diffusion coefficient, which decreases\\
\emph{GPT-5.5 answers.} Apparent diffusion coefficient decreases; Apparent diffusion coefficient; decreased in active inflammation; Apparent diffusion coefficient decreases; Apparent diffusion coefficient; decreased ADC; Decreased apparent diffusion coefficient; Decreased apparent diffusion coefficient; Apparent diffusion coefficient; decreased ADC; Apparent diffusion coefficient; decreased in active inflammation\end{quote}

\paragraph{GPT-5.5.}
A multiple-choice task the student answers correctly in all eight rollouts. The reference key is A.
\begin{quote}\small A 26-year-old fit man is newly diagnosed with Philadelphia chromosome-negative precursor T-cell acute lymphoblastic leukemia/lymphoblastic lymphoma with marrow involvement and no CNS disease. The team is choosing initial therapy and wants to avoid unnecessary hospitalization and late toxicities while preserving disease control. Which management plan is most appropriate?
\begin{itemize}\itemsep0pt
\item[A.] Use a pediatric-inspired, asparaginase-containing ALL regimen with CNS prophylaxis rather than defaulting to adult hyper-CVAD
\item[B.] Use CHOP because lymphoblastic lymphoma is managed like other aggressive non-Hodgkin lymphomas
\item[C.] Use hyper-CVAD induction and plan allogeneic hematopoietic cell transplantation in first complete remission for all patients regardless of risk or response
\item[D.] Use vincristine and prednisone alone for induction to reduce early infectious mortality in young adults
\end{itemize}\end{quote}
A short-answer task with mixed student correctness has 5 of 8 answers accepted by the reference grader.
\begin{quote}\small A diabetic patient with severe peripheral vascular disease undergoes transmetatarsal amputation for forefoot gangrene, followed by infrainguinal bypass using a mapped patent great saphenous vein. Postoperatively the bypass remains patent, blood cultures are negative, and chest imaging shows no pneumonia, but the amputation stump dehisces with progressive necrosis and failure to heal despite wound VAC and antibiotics. What vascular problem can explain limb-salvage failure despite a patent bypass graft?\\[2pt]
\emph{Reference answer.} poor distal runoff\\
\emph{Student answers.} Inadequate distal runoff; inadequate distal runoff; Distal microembolization; Inadequate distal runoff; Diabetic microvascular disease; Diabetic microangiopathy; poor distal runoff; Inadequate distal runoff\\
\emph{GPT-5.5 answers.} poor pedal arch perfusion; poor distal pedal runoff; poor pedal arch outflow; occluded pedal arch; severe pedal small-vessel disease; poor pedal arch outflow; severe inframalleolar occlusive disease; severe inframalleolar occlusive disease\end{quote}

\paragraph{Qwen3.5-397B.}
A multiple-choice task the student answers correctly in all eight rollouts. The reference key is B.
\begin{quote}\small A 68-year-old female with type 2 diabetes mellitus and stage 3 chronic kidney disease with an eGFR of 45 mL/min presents with altered mental status. She has been on chronic metformin therapy. Laboratory evaluation reveals a pH of 7.10, bicarbonate of 12 mEq/L, an anion gap of 24 mEq/L, and a lactate level of 9 mmol/L. Ketones are negative. Based on the provided medical literature, what is the primary pathophysiological mechanism responsible for this patient's metabolic derangement?
\begin{itemize}\itemsep0pt
\item[A.] Increased lipolysis leading to excessive beta-hydroxybutyrate accumulation
\item[B.] Inhibition of mitochondrial respiratory chain complex I and hepatic gluconeogenesis
\item[C.] Impaired renal excretion of endogenous organic acids due to low GFR alone
\item[D.] Direct stimulation of anaerobic glycolysis in skeletal muscle due to hypoxia
\end{itemize}\end{quote}
A short-answer task with mixed student correctness has 6 of 8 answers accepted by the reference grader.
\begin{quote}\small In an immunocompromised patient with myelodysplastic syndrome and fever, chest CT reveals multiple small peribronchial nodules under 5 mm without cavitation. Bronchoalveolar lavage isolates Parainfluenza type 3, while tuberculosis and fungal studies are negative. Based on literature describing this virus's specific radiographic signature in immunocompromised hosts distinct from miliary tuberculosis or Pneumocystis pneumonia, what is the diagnosis?\\[2pt]
\emph{Reference answer.} Parainfluenza virus type 3 pneumonia\\
\emph{Student answers.} Parainfluenza type 3 pneumonia; suspicious for respiratory syncytial virus infection; Parainfluenza type 3 pneumonia; Parainfluenza virus 3 pneumonitis; human parainfluenza virus 3 pneumonia; Parainfluenza type 3 pneumonitis; Respiratory syncytial virus pneumonia; Parainfluenzal pneumonitis\\
\emph{GPT-5.5 answers.} Parainfluenza type 3 pneumonia; Parainfluenza type 3 pneumonia; Parainfluenza type 3 pneumonia; Parainfluenza type 3 pneumonia; Parainfluenza type 3 pneumonia; Parainfluenza type 3 pneumonia; Parainfluenza type 3 pneumonia; Parainfluenza type 3 pneumonia\end{quote}

\paragraph{DeepSeek-V4-Pro.}
A multiple-choice task the student answers correctly in all eight rollouts. The reference key is B.
\begin{quote}\small A 52-year-old man with type 2 diabetes and end-stage renal disease is about to start dialysis. He is relatively young and otherwise stable. According to observational studies comparing dialysis modalities in diabetic patients, which modality has been associated with equal or better survival during the first few years of treatment for younger diabetic individuals?
\begin{itemize}\itemsep0pt
\item[A.] In-center hemodialysis
\item[B.] Peritoneal dialysis
\item[C.] Both modalities offer identical survival regardless of age
\item[D.] Immediate preemptive renal transplantation
\end{itemize}\end{quote}
A short-answer task with mixed student correctness has 4 of 8 answers accepted by the reference grader.
\begin{quote}\small Which cardiomyopathy can present on an electrocardiogram with transient J waves that appear before the development of ST-segment elevation or T wave inversion?\\[2pt]
\emph{Reference answer.} Takotsubo cardiomyopathy\\
\emph{Student answers.} Arrhythmogenic right ventricular cardiomyopathy; Brugada syndrome; no boxed answer; Takotsubo cardiomyopathy; Takotsubo cardiomyopathy; no boxed answer; Takotsubo cardiomyopathy; Takotsubo cardiomyopathy\\
\emph{GPT-5.5 answers.} Takotsubo cardiomyopathy; Takotsubo cardiomyopathy; Takotsubo cardiomyopathy; Takotsubo cardiomyopathy; Takotsubo cardiomyopathy; Takotsubo cardiomyopathy; Takotsubo cardiomyopathy; Takotsubo cardiomyopathy\end{quote}

\paragraph{Kimi-K2.6.}
A multiple-choice task the student answers correctly in all eight rollouts. The reference key is B.
\begin{quote}\small A 12-year-old boy with new-onset type 1 diabetes presents with severe DKA with pH 6.92, serum bicarbonate 5 mmol/L, and PaCO2 8 mmHg. His parents report he drank several liters of juice and water over the preceding 12 hours. Thirty minutes after initiating isotonic crystalloid resuscitation and an insulin infusion, he becomes obtunded with sluggish pupillary responses. Which characterization and immediate management step is most consistent with the pathophysiology and risk factors of this complication?
\begin{itemize}\itemsep0pt
\item[A.] Late-onset cerebral edema caused by rapid insulin-mediated correction of serum osmolarity; slow the insulin infusion rate.
\item[B.] Early-onset cerebral edema associated with pre-hospital fluid intake and severe hypocapnia; administer mannitol and consider hyperventilation.
\item[C.] Excessive parenteral fluid administration causing dilutional hyponatremia and brain swelling; restrict fluids and give furosemide.
\item[D.] Progression of severe metabolic acidosis causing cerebral hypoperfusion; administer sodium bicarbonate and increase the fluid rate.
\end{itemize}\end{quote}
A short-answer task with mixed student correctness has 5 of 8 answers accepted by the reference grader.
\begin{quote}\small A patient from a Lyme-endemic area has intermittent knee monoarthritis with negative RF, ANA, and blood/synovial cultures, and negative serum Lyme serology by two-tier testing, but synovial fluid PCR is positive for Borrelia DNA. What specific subtype of Lyme arthritis is most likely?\\[2pt]
\emph{Reference answer.} Seronegative Lyme arthritis\\
\emph{Student answers.} Seronegative Lyme arthritis; Infection-driven Lyme arthritis; Destructive, necrotizing Lyme arthritis; Seronegative Lyme arthritis; seronegative Lyme arthritis; Seronegative Lyme arthritis; Recurrent monoarticular Lyme arthritis; Early Lyme arthritis\\
\emph{GPT-5.5 answers.} seronegative Lyme arthritis; seronegative Lyme arthritis; seronegative Lyme arthritis; seronegative Lyme arthritis; seronegative Lyme arthritis; seronegative Lyme arthritis; seronegative Lyme arthritis; seronegative Lyme arthritis\end{quote}

\paragraph{GPT-5.5 answering its own task.} GPT-5.5 gives equivalent correct answers in all eight samples; the student is correct in six.
\begin{quote}\small A 67-year-old man is evaluated after an incidental monoclonal IgG kappa spike is found. Bone marrow biopsy shows 18\% CD138-positive kappa-restricted plasma cells. Hemoglobin, creatinine, calcium, and skeletal survey are normal, and he has no bone pain. Whole-body MRI, however, shows two separate 6-mm focal marrow lesions. Under current IMWG criteria, what diagnosis should be made rather than MGUS or smoldering myeloma?\\[2pt]\emph{Reference answer.} Multiple myeloma\\\emph{GPT-5.5 answers.} multiple myeloma; active multiple myeloma; multiple myeloma; active multiple myeloma; multiple myeloma; multiple myeloma; active multiple myeloma; active multiple myeloma\\\emph{Student answers.} multiple myeloma; active multiple myeloma; Multiple myeloma; Multiple myeloma; active multiple myeloma; none; Multiple myeloma; none\end{quote}

\paragraph{A student-generated task with mixed correctness.} GPT-5.5 answers correctly in all eight samples; the student in four.
\begin{quote}\small In a patient with hepatocellular carcinoma being evaluated for transcatheter arterial chemoembolization, complete portal vein thrombosis is traditionally considered a major contraindication. What catastrophic hepatic complication does this precaution specifically aim to prevent?\\[2pt]\emph{Reference answer.} Massive necrosis of the non-tumoral liver\\\emph{Student answers.} acute hepatic failure; acute hepatic failure; massive hepatic necrosis; Hepatic infarction; acute liver failure; massive hepatic necrosis; acute liver failure; massive hepatic necrosis\end{quote}

%% file: sections/appendix_examples.tex
\section{Example generated tasks}
\label{app:examples}

The following tasks and retrieval queries are reproduced verbatim from the proposer's pool.
They illustrate questions generated about clinical mechanisms and treatment decisions.

\paragraph{Example 1.}
\emph{Retrieval query.} ``What is the pathophysiological basis for the NIH Stroke Scale's
differential weighting of left versus right hemisphere stroke deficits?''

\emph{Generated question.} ``In patients with right hemisphere stroke, how might the NIHSS
scoring system impact clinical trial eligibility if a low score is used as an exclusion
criterion?''
\begin{itemize}\itemsep0pt
\item[A.] The NIHSS may underrepresent right hemisphere deficits, leading to under-enrollment
of these patients. \textbf{Correct answer.}
\item[B.] The NIHSS is equally sensitive to both hemispheres and does not favor one side.
\item[C.] The NIHSS penalizes right hemisphere strokes more heavily due to greater motor
impairment.
\item[D.] Right hemisphere strokes are always excluded because they have higher NIHSS scores
than left hemisphere strokes.
\end{itemize}
The NIH Stroke Scale (NIHSS) summarizes stroke deficits. This example tests whether the model connects the scale's weighting to who qualifies for a trial. The retrieved passages support the reference answer by describing the allocation of seven points to language and two to neglect, which is a failure to attend to one side of space.

\paragraph{Example 2.}
\emph{Retrieval query.} ``What is the pathophysiological rationale for using vasopressors
during aortic perfusion and oxygenation in cardiac arrest patients?''

\emph{Generated question.} ``In a patient undergoing aortic balloon occlusion during CPR, which
strategy is most likely to enhance myocardial perfusion while minimizing adverse effects on
cerebral blood flow, based on hemodynamic principles and pharmacologic profiles?''
\begin{itemize}\itemsep0pt
\item[A.] Intravenous vasopressin administration
\item[B.] Intra-aortic vasopressin delivery above the balloon occlusion \textbf{Correct answer.}
\item[C.] Nitroglycerin infusion to reduce afterload
\item[D.] High-dose epinephrine to increase contractility
\end{itemize}
The answer choices describe interventions drawn from the retrieved evidence. This task asks the model to connect drug delivery and blood flow. Myocardial perfusion means blood reaching the heart muscle, while cerebral blood flow supplies the brain.

%% file: sections/appendix_extended_results.tex
\section{Detailed results and analysis}
\label{sec:results}

\subsection{Compute Scaling and Inference Cost}
\label{sec:scaling}

Figure~\ref{fig:teaser}(b) shows that continuous task generation supports sustained improvement
with training compute. The 27B runs share supervised initialization, reward, judge, and optimizer,
and their compute covers proposal, rollouts, judging, and policy updates, as detailed in
Appendix~\ref{app:training}. The 27B model increases from $0.421$ to $0.516$ accuracy at
approximately $3.9$ points per $10^{20}$ FLOPs, with no observed saturation. A linear fit
explains $96\%$ of the variance, compared with $83\%$ for a log-linear fit. The run includes $197$
validation passes over all $2452$ cases.

RL on the labeled training set peaks at $0.463$ and then collapses to $0.133$. Training on the
static pool peaks at $0.441$ and declines to $0.359$ over the final thirty
steps. The comparison with these two fixed task sets shows the benefit of generating
additional tasks as training progresses.

The 9B model keeps improving after $5\times10^{19}$ FLOPs, from about $0.46$ to $0.488$ at its last checkpoint after $1.4\times10^{20}$ FLOPs. Its training steps cost approximately one third as much as those
of the 27B model, allowing it to exceed RSIMed-27B and the static pool at matched compute.

\textbf{Even the cheapest GPT-5.6 setting costs twice as much as RSIMed-27B and is less accurate.}
\label{sec:cost}
Figure~\ref{fig:teaser}(a) compares diagnostic accuracy with the estimated cost of answering all
$2452$ cases. The 27B and 9B models cost \$0.57 and \$0.17, respectively, using the serving
rates of Table~\ref{tab:pricing} and measured response lengths of $383$ and $268$ output tokens per case.
We scale the 9B serving rate by its relative per-token compute, $0.35\times$ the 27B rate. RSIMed-27B writes about 380 output tokens per case, so reading the admission record and
writing the answer account for roughly equal shares of its cost. Even the cheapest GPT-5.6
setting, Luna without reasoning, costs twice as much as RSIMed-27B and reaches 0.326. From no
reasoning to the highest effort, accuracy rises by 7.3 points for Luna, 4.8 points for Terra,
and 5.3 points for Sol, while reasoning tokens make up 77\% to 98\% of the billed output at
every setting that reasons.

\subsection{Diagnostic Performance}
\label{sec:vs-trainset}

\begin{figure*}[t]
\centering
\includegraphics[width=\textwidth]{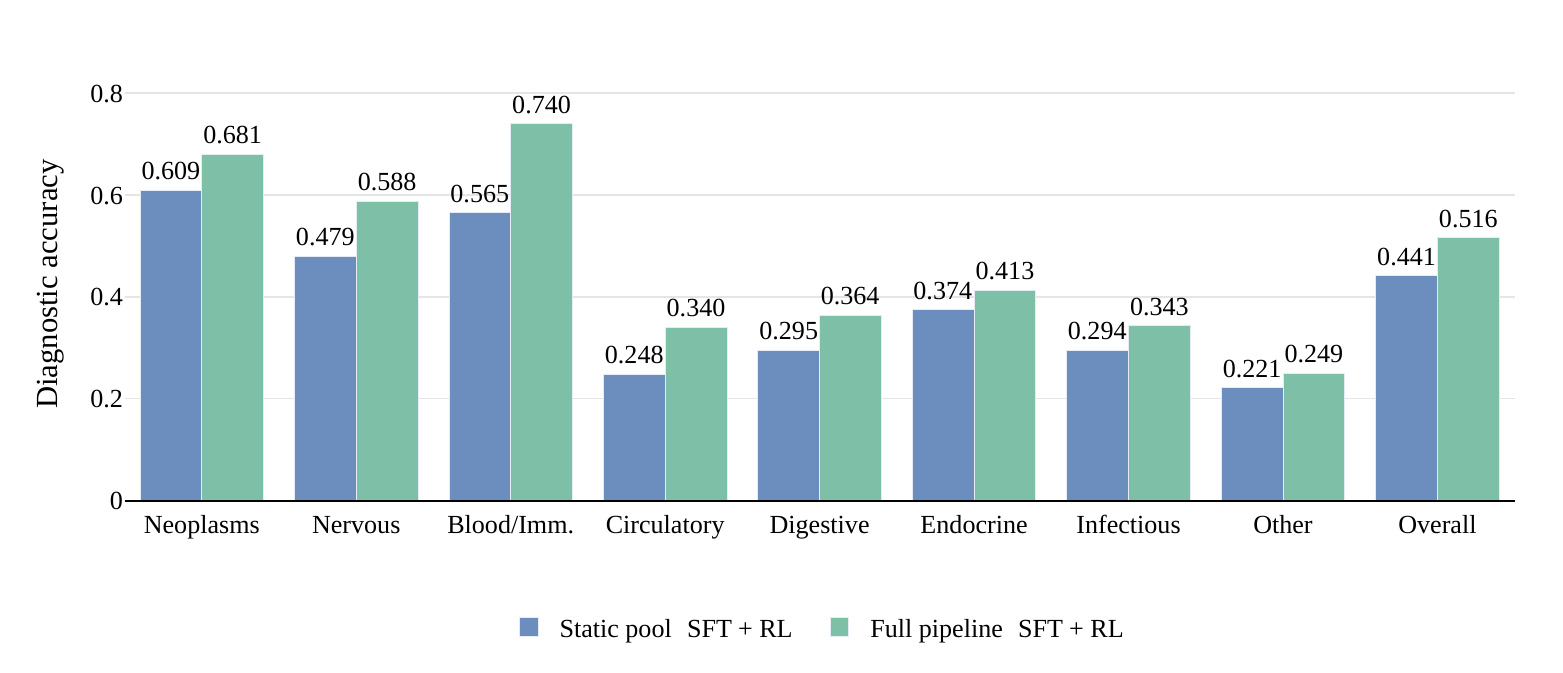}
\caption{\textbf{Continuous generation versus a static task pool.} The full 27B pipeline improves full-set MIMIC-IV accuracy in every ICD chapter and overall from $0.441$ to $0.516$ relative to a frozen $15{,}000$-task pool, with both runs initialized by supervised fine-tuning and scored by the same fixed judge.}
\label{fig:delta}
\end{figure*}

Table~\ref{tab:main} reports the full pipeline's diagnostic performance. The two-stage
27B model reaches $0.516$ overall accuracy and improves over its base in every
ICD chapter. The largest gains include blood and immune disorders, where accuracy rises
from the base model's $0.332$ to $0.740$, and endocrine and metabolic diseases, where it
rises from $0.164$ to $0.413$.

Among frozen-backbone inference methods, self-consistency achieves the highest
overall accuracy, at $0.272$ for 9B and $0.346$ for 27B, compared with matched direct
scores of $0.262$ and $0.332$. The shared-index RAG adaptations score below the
matched direct evaluation on both backbones. The 27B static-pool training control
reaches $0.441$, while the full pipeline reaches $0.516$.

Figure~\ref{fig:delta} compares the full pipeline with static-pool training.
Continuous generation achieves higher accuracy in every ICD chapter and improves overall
accuracy by $7.5$ points. The largest chapter gain is in blood and immune disorders,
from $0.565$ to $0.740$, and nervous-system accuracy rises from $0.479$ to $0.588$ and
circulatory accuracy from $0.248$ to $0.340$.

\input{tables/main_results}
\input{tables/healthbench_results}

\subsection{What the RL Stage Adds}
\label{app:rl-stage}
\input{tables/stage_ablation}
Table~\ref{tab:stages} gives the scores of each training stage on both benchmarks. RL supplies 52\% of the total MIMIC-IV gain of the 27B model and 88\% of that of the 9B model, whose warm start moves accuracy only from 0.261 to 0.288 before RL lifts it to 0.488. We regrade the supervised checkpoint and the final checkpoint case by case with the same fixed
judge. The RL stage, which trains on generated tasks mixed with the labeled cases, solves 329
evaluation cases that the supervised checkpoint missed and loses 98 that it had solved, a net
gain of 231 of the 2,452 cases. The gains appear in every ICD chapter and are largest for
nervous-system disease, neoplasms, and blood and immune disorders, with net gains of 52, 51,
and 50 cases.

\subsection{Out-of-Distribution Clinical Performance}
\label{sec:healthbench}

Table~\ref{tab:main-hb} evaluates out-of-distribution transfer from rare-disease diagnosis training to the distinct consultation, clinical writing, and research tasks in HealthBench Professional. Despite this shift in task types, diagnostic training improves performance on HealthBench Professional. RSIMed-RL-27B scores $0.303$ length-adjusted and $0.376$ raw, exceeding its base model
by $1.2$ and $1.3$ points and outperforming the evaluated open medical models on both
metrics. Among the open base models, Qwen3.5-9B scores $0.187$ and Gemma-4-31B $0.293$
length-adjusted, and among the stronger models, Kimi-K2.6 scores $0.335$ and Grok-4.3 $0.415$.
Appendix~\ref{app:results} reports the RL-only transfer breakdown.

\textbf{The trained model improves on adversarial prompts without safety-specific training.} HealthBench Professional includes 256 ordinary clinical requests in the \emph{good-faith typical} slice, 78 difficult legitimate requests in the \emph{good-faith difficult} slice, and 191 adversarial prompts intended to elicit unsafe or unprofessional responses in the \emph{red-teaming difficult} slice. The adversarial rubrics assess refusals, redirection,
and explicit identification of risks. Table~\ref{tab:hb-slices} reports these slices graded by
GPT-5.5-Chat, without safety-specific training or prompting.
On adversarial prompts, the self-trained 27B model scores $0.047$, compared with $0.030$ for
its base.
The 27B model also improves on both good-faith slices, and difficult good-faith requests rise
from $0.174$ to $0.212$. By use case, the raw score rises by $0.017$ on consultation, $0.018$ on
writing, and $0.003$ on research prompts.

\subsection{Analysis of Teacher Quality}
\label{sec:why}

\begin{figure*}[t]
\centering
\includegraphics[width=0.62\textwidth]{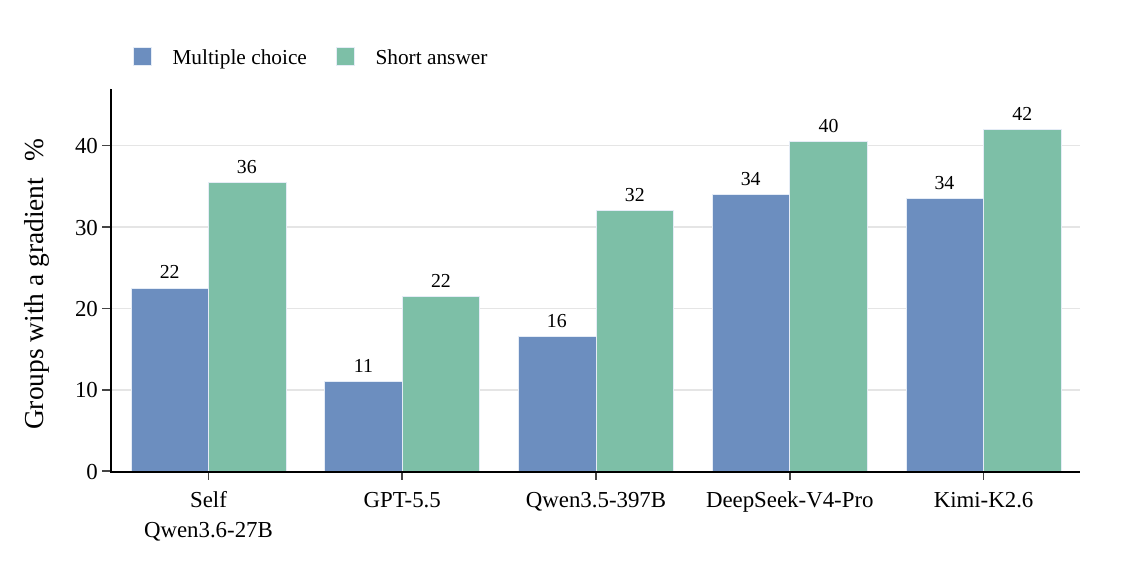}
\caption{\textbf{Correctness-reward variation by task source.} Self-generated tasks yield more groups with unequal rewards than GPT-5.5's tasks when the untrained student answers each task eight times with training-time sampling and a common correctness scorer, providing a proxy for group-relative learning opportunities under the protocol in Appendix~\ref{app:teacher}.}
\label{fig:teacher-signal}
\end{figure*}

\textbf{Self-generated tasks give the student more groups with mixed rewards than GPT-5.5's tasks.} We sample eight answers per task from the
untrained student and use a common correctness scorer to measure variation within each
group. Figure~\ref{fig:teacher-signal} shows that self-generated tasks produce varying
correctness rewards in $22\%$ of multiple-choice groups and $36\%$ of short-answer groups,
compared with $11\%$ and $22\%$ for GPT-5.5's tasks. Such groups provide opportunities
for group-relative learning under this correctness reward.

\textbf{Retrieved passages help the student more on self-generated tasks than on GPT-5.5's tasks.} We supply the proposer's retrieved passages to
the student to measure how much they help it answer each task. On self-generated short-answer
tasks, accuracy increases from $0.787$ to $0.901$, compared with $0.925$ to $0.979$ on
GPT-5.5's tasks. The larger improvement on self-generated tasks suggests that their passages
contain information the student needs to answer correctly. GPT-5.5 answers the student's
multiple-choice tasks with $0.978$ accuracy, as reported in Table~\ref{tab:solver-matrix}.

\begin{figure*}[t]
\centering
\includegraphics[width=\textwidth]{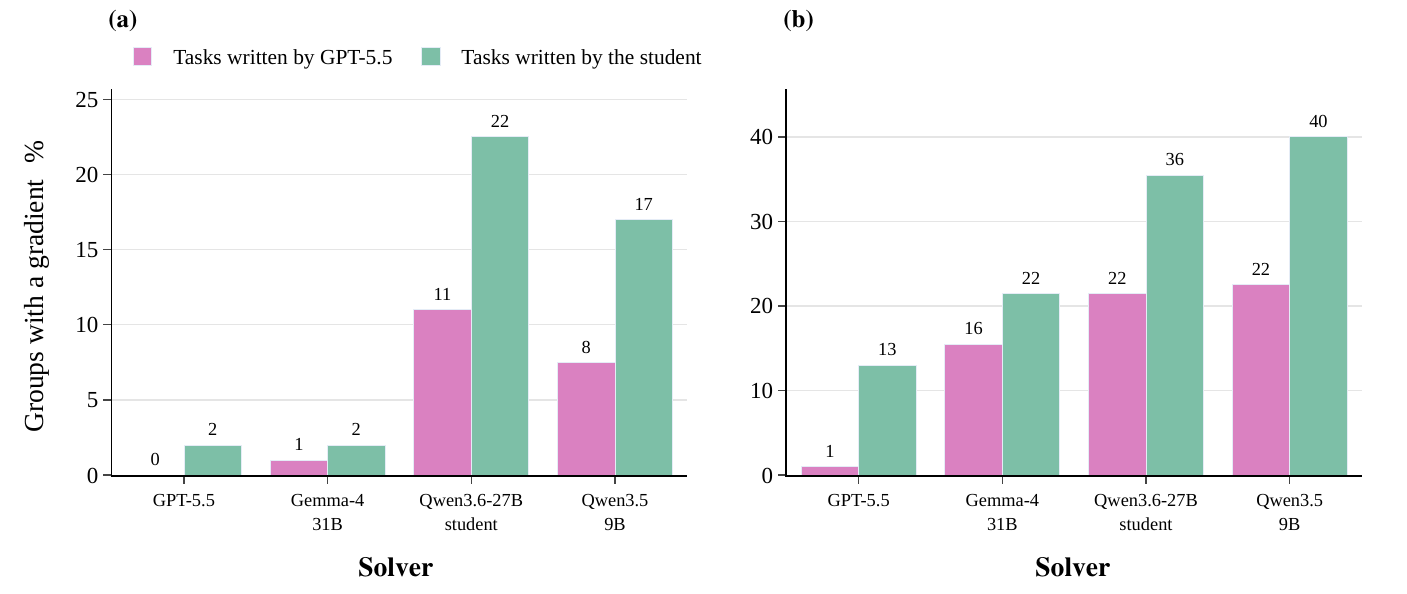}
\caption{\textbf{(a) Multiple-choice task difficulty.} Self-generated tasks produce more groups with varying correctness rewards than GPT-5.5's tasks across all four solvers sampled eight times per task.
\textbf{(b) Short-answer task difficulty.} The same pattern holds for short-answer tasks; Table~\ref{tab:solver-matrix} reports accuracy and the subset GPT-5.5 answers correctly in all eight samples.}
\label{fig:solver-matrix}
\end{figure*}

\textbf{Other solvers also find self-generated tasks harder than GPT-5.5's tasks.} Figure~\ref{fig:solver-matrix} shows that the difficulty
difference extends to Gemma-4-31B and Qwen3.5-9B. Both produce more groups with varying
correctness rewards on the student's tasks than on GPT-5.5's. On the expanded multiple-choice
subset that GPT-5.5 answers correctly in all eight samples, self-generated tasks reduce
accuracy by $1.5$ points for Gemma and $1.8$ points for the student relative to GPT-5.5's
tasks.

%% file: tables/main_results.tex
\begin{table*}[t]
\caption{MIMIC-IV rare-disease diagnosis on the 2,452 evaluation cases, overall and by ICD chapter, scored with a fixed
judge, as detailed in Appendix~\ref{app:judging}. Chapter sizes are Neopl.\ $695$, Endo.\ $494$, Nerv.\ $388$, Blood $223$, Circ.\ $141$,
Infect.\ $102$, Digest.\ $88$, Other $321$.}
\label{tab:main}
\begin{center}
\footnotesize
\setlength{\tabcolsep}{3pt}
\resizebox{\textwidth}{!}{%
\begin{tabular}{lccccccccc}
\toprule
& \multicolumn{9}{c}{MIMIC-IV rare disease} \\
\cmidrule(lr){2-10}
Model & Neopl. & Endo. & Nerv. & Blood & Circ. & Infect. & Digest. & Other & Overall \\
\midrule
\multicolumn{10}{l}{\emph{Open medical models}} \\
\quad MedGemma-1.5-4B & 0.227 & 0.121 & 0.070 & 0.175 & 0.021 & 0.147 & 0.125 & 0.028 & 0.131 \\
\quad Lingshu-7B & 0.249 & 0.087 & 0.149 & 0.224 & 0.099 & 0.118 & 0.170 & 0.044 & 0.155 \\
\quad Lingshu-32B & 0.420 & 0.152 & 0.216 & 0.242 & 0.113 & 0.186 & 0.159 & 0.081 & 0.237 \\
\midrule
\multicolumn{10}{l}{\emph{Untrained base models}} \\
\quad Qwen3.5-9B & 0.452 & 0.079 & 0.325 & 0.269 & 0.156 & 0.225 & 0.205 & 0.118 & 0.261 \\
\quad Gemma-4-31B & 0.468 & 0.097 & 0.302 & 0.336 & 0.149 & 0.206 & 0.295 & 0.128 & 0.275 \\
\quad Qwen3.6-27B & 0.548 & 0.164 & 0.389 & 0.332 & 0.156 & 0.294 & 0.261 & 0.178 & 0.334 \\
\midrule
\multicolumn{10}{l}{\emph{Larger reference models}} \\
\quad Qwen3.5-397B & 0.540 & 0.119 & 0.351 & 0.238 & 0.135 & 0.275 & 0.318 & 0.131 & 0.302 \\
\midrule
\multicolumn{10}{l}{\emph{9B curated-data control (SFT + RL, self-judge)}} \\
Qwen3.5-9B & 0.583 & 0.360 & 0.472 & 0.498 & 0.199 & 0.225 & 0.330 & 0.171 & 0.413 \\
\midrule
\multicolumn{10}{l}{\emph{Frozen Qwen3.5-9B inference methods}} \\
\quad Direct (matched run) & 0.450 & 0.079 & 0.332 & 0.287 & 0.142 & 0.235 & 0.193 & 0.112 & 0.262 \\
\quad Medical RAG & 0.422 & 0.126 & 0.232 & 0.247 & 0.121 & 0.196 & 0.284 & 0.121 & 0.245 \\
\quad RAG-Fusion & 0.403 & 0.099 & 0.232 & 0.278 & 0.121 & 0.245 & 0.227 & 0.125 & 0.238 \\
\quad i-MedRAG & 0.393 & 0.087 & 0.214 & 0.283 & 0.099 & 0.235 & 0.216 & 0.112 & 0.226 \\
\quad Self-consistency (5 samples) & 0.462 & 0.111 & 0.343 & 0.265 & 0.142 & 0.216 & 0.239 & 0.112 & 0.272 \\
\quad Self-refinement & 0.384 & 0.061 & 0.265 & 0.269 & 0.142 & 0.186 & 0.136 & 0.093 & 0.221 \\
\quad Chain-of-verification & 0.391 & 0.065 & 0.276 & 0.283 & 0.156 & 0.186 & 0.227 & 0.090 & 0.230 \\
\midrule
\multicolumn{10}{l}{\emph{Frozen Qwen3.6-27B inference methods}} \\
\quad Direct (matched run) & 0.540 & 0.160 & 0.387 & 0.354 & 0.156 & 0.265 & 0.273 & 0.181 & 0.332 \\
\quad Medical RAG & 0.534 & 0.196 & 0.343 & 0.341 & 0.156 & 0.294 & 0.273 & 0.168 & 0.329 \\
\quad RAG-Fusion & 0.511 & 0.162 & 0.348 & 0.390 & 0.113 & 0.255 & 0.261 & 0.181 & 0.318 \\
\quad i-MedRAG & 0.515 & 0.150 & 0.309 & 0.327 & 0.121 & 0.304 & 0.193 & 0.153 & 0.301 \\
\quad Self-consistency (5 samples) & 0.584 & 0.164 & 0.407 & 0.345 & 0.170 & 0.284 & 0.295 & 0.150 & 0.346 \\
\quad Self-refinement & 0.465 & 0.123 & 0.273 & 0.327 & 0.128 & 0.275 & 0.273 & 0.134 & 0.276 \\
\quad Chain-of-verification & 0.545 & 0.150 & 0.392 & 0.332 & 0.170 & 0.304 & 0.318 & 0.184 & 0.335 \\
\midrule
\multicolumn{10}{l}{\emph{Qwen3.6-27B training controls}} \\
\quad SFT + train-set RL & 0.635 & 0.464 & 0.487 & 0.529 & 0.259 & 0.286 & 0.266 & 0.213 & 0.463 \\
\quad SFT + static-pool RL & 0.609 & 0.374 & 0.479 & 0.565 & 0.248 & 0.294 & 0.295 & 0.221 & 0.441 \\
\midrule
\method{}-9B (ours) & 0.607 & \textbf{0.518} & 0.503 & 0.731 & \textbf{0.348} & 0.324 & 0.307 & 0.159 & 0.488 \\
\textbf{\method{}-27B (ours)} & \textbf{0.681} & 0.413 & \textbf{0.588} & \textbf{0.740} & 0.340 & \textbf{0.343} & \textbf{0.364} & \textbf{0.249} & \textbf{0.516} \\
\bottomrule
\end{tabular}%
}
\end{center}
\end{table*}

%% file: tables/healthbench_results.tex
\begin{table*}[t]
\caption{HealthBench Professional on its 525 tasks, graded by GPT-5.5-Chat. Columns show the
length-adjusted and raw rubric scores. Our row uses the RL-only variant evaluated on this
benchmark, as detailed in Appendix~\ref{app:variant}.}
\label{tab:main-hb}
\begin{center}
\footnotesize
\begin{tabular}{lcc}
\toprule
Model & Len-adj. & Raw \\
\midrule
\multicolumn{3}{l}{\emph{Open medical models}} \\
\quad MedGemma-1.5-4B & 0.069 & 0.161 \\
\quad Lingshu-7B & 0.061 & 0.061 \\
\quad Lingshu-32B & 0.145 & 0.130 \\
\midrule
\multicolumn{3}{l}{\emph{Untrained base models}} \\
\quad Qwen3.5-9B & 0.187 & 0.255 \\
\quad Gemma-4-31B & 0.293 & 0.326 \\
\quad Qwen3.6-27B & 0.291 & 0.363 \\
\quad DeepSeek-V4-Pro & 0.282 & 0.360 \\
\midrule
\textbf{RSIMed-RL-27B (ours)} & 0.303 & 0.376 \\
\bottomrule
\end{tabular}
\end{center}
\end{table*}

%% file: tables/stage_ablation.tex
\begin{table}[t]
\caption{\textbf{Contribution of each training stage for Qwen3.6-27B.} The RL-only run trains on
self-generated tasks from the base weights, and the full pipeline runs RL after the supervised
warm start. Both benchmarks are graded by GPT-5.5-Chat.}
\label{tab:stages}
\begin{center}
\small
\begin{tabular}{lccc}
\toprule
& MIMIC-IV & \multicolumn{2}{c}{HealthBench Prof.} \\
\cmidrule(lr){2-2}\cmidrule(lr){3-4}
Training & Accuracy & Len-adj. & Raw \\
\midrule
Untrained & 0.334 & 0.291 & 0.363 \\
Supervised warm start only & 0.421 & 0.290 & 0.368 \\
RL only & - & \textbf{0.303} & \textbf{0.376} \\
Warm start and RL, full pipeline & \textbf{0.516} & 0.280 & 0.349 \\
\bottomrule
\end{tabular}
\end{center}
\end{table}

%% file: sections/appendix_limitations.tex
\section{Limitations}
\label{app:limitations}

The evaluation covers rare-disease diagnosis and transfer to HealthBench Professional, and broader task coverage would help assess generalization. Model-generated tasks, reference answers, and model-based grading can contain errors despite retrieval and validation. Inference costs are estimates under the stated serving-rate assumptions. The reported benchmark scores measure model performance on these tasks rather than patient outcomes or clinical deployment effectiveness.

%% file: sections/appendix_availability.tex
\section{Data availability and ethics}
\label{app:availability}

\paragraph*{Data and Code Availability}
This work evaluates models on MIMIC-IV, a public dataset with credentialed access through PhysioNet, and on the public HealthBench Professional benchmark. The medical database includes additional public datasets described in Section~\ref{sec:method}. All code, prompts, and configurations are released at \url{https://github.com/DDVD233/RRI}.

\paragraph*{Institutional Review Board}
No new human subjects research was conducted. All data are reused from publicly available, pre-anonymized clinical datasets that obtained their own ethical approvals at the time of original release.